\documentclass{article}

\usepackage{nac_preprint}

\usepackage[utf8]{inputenc} 
\usepackage[T1]{fontenc}    
\usepackage{hyperref}       
\usepackage{url}            
\usepackage{booktabs}       
\usepackage{amsfonts}       
\usepackage{nicefrac}       
\usepackage{microtype}      

\usepackage{booktabs} 
\usepackage{graphicx,subcaption,float}
\usepackage{amsmath}
\usepackage{mathtools}
\usepackage[toc,page]{appendix}
\usepackage{enumitem}
\usepackage{graphics}
\usepackage{graphicx}
\usepackage{algorithm}
\usepackage{algorithmicx}
\usepackage[noend]{algpseudocode}
\algnewcommand{\LineComment}[1]{\State \(//\) #1}
\usepackage{enumitem}
\usepackage{xr}

\usepackage{etoolbox}
\usepackage{tikz}
\usetikzlibrary{tikzmark}
\usetikzlibrary{calc}

\errorcontextlines\maxdimen

\newcommand{\ALGtikzmarkcolor}{black}
\newcommand{\ALGtikzmarkextraindent}{4pt}
\newcommand{\ALGtikzmarkverticaloffsetstart}{-.5ex}
\newcommand{\ALGtikzmarkverticaloffsetend}{-.5ex}

\makeatletter
\newcommand*{\addFileDependency}[1]{
  \typeout{(#1)}
  \@addtofilelist{#1}
  \IfFileExists{#1}{}{\typeout{No file #1.}}
}
\makeatother

\makeatletter
\newcounter{ALG@tikzmark@tempcnta}

\newcommand\ALG@tikzmark@start{%
    \global\let\ALG@tikzmark@last\ALG@tikzmark@starttext%
    \expandafter\edef\csname ALG@tikzmark@\theALG@nested\endcsname{\theALG@tikzmark@tempcnta}%
    \tikzmark{ALG@tikzmark@start@\csname ALG@tikzmark@\theALG@nested\endcsname}%
    \addtocounter{ALG@tikzmark@tempcnta}{1}%
}

\def\ALG@tikzmark@starttext{start}
\newcommand\ALG@tikzmark@end{%
    \ifx\ALG@tikzmark@last\ALG@tikzmark@starttext
    \else
        \tikzmark{ALG@tikzmark@end@\csname ALG@tikzmark@\theALG@nested\endcsname}%
        \tikz[overlay,remember picture] \draw[\ALGtikzmarkcolor] let \p{S}=($(pic cs:ALG@tikzmark@start@\csname ALG@tikzmark@\theALG@nested\endcsname)+(\ALGtikzmarkextraindent,\ALGtikzmarkverticaloffsetstart)$), \p{E}=($(pic cs:ALG@tikzmark@end@\csname ALG@tikzmark@\theALG@nested\endcsname)+(\ALGtikzmarkextraindent,\ALGtikzmarkverticaloffsetend)$) in (\x{S},\y{S})--(\x{S},\y{E});%
    \fi
    \gdef\ALG@tikzmark@last{end}%
}

\apptocmd{\ALG@beginblock}{\ALG@tikzmark@start}{}{\errmessage{failed to patch}}
\pretocmd{\ALG@endblock}{\ALG@tikzmark@end}{}{\errmessage{failed to patch}}
\makeatother

\title{Active Predicting Coding: Brain-Inspired Reinforcement Learning for Sparse Reward Robotic Control Problems}

\author{%
  Alexander Ororbia \\
  Rochester Institute of Technology \\
  \texttt{ago@cs.rit.edu}
  \And 
  Ankur Mali \\
  The University of South Florida \\
  \texttt{ankurarjunmali@usf.edu}
}

\begin{document}
\setlength{\abovedisplayskip}{0.065cm}
\setlength{\belowdisplayskip}{0pt}

\maketitle

\begin{abstract}

In this article, we propose a backpropagation-free approach to robotic control through the neuro-cognitive computational framework of neural generative coding (NGC), designing an agent built completely from powerful predictive coding/processing circuits that facilitate dynamic, online learning from sparse rewards, embodying the principles of planning-as-inference.
Concretely, we craft an adaptive agent system, which we call active predictive coding (ActPC), that balances an internally-generated epistemic signal (meant to encourage intelligent exploration) with an internally-generated instrumental signal (meant to encourage goal-seeking behavior) to ultimately learn how to control various simulated robotic systems as well as a complex robotic arm using a realistic robotics simulator, i.e., the Surreal Robotics Suite, for the block lifting task and can pick-and-place problems.
Notably, our experimental results demonstrate that our proposed ActPC agent performs well in the face of sparse (extrinsic) reward signals and is competitive with or outperforms several powerful backprop-based RL approaches.

\end{abstract}


\section{Introduction} 
\label{sec:intro}

One of the key goals of brain-inspired computing is to develop methods that draw inspiration from computational neuroscience and cognitive science to build effective  adaptive and efficient agents that are capable of intelligently interacting with their environment. Notably, brain-inspired computational research seeks to develop intelligent systems that are capable of circumventing the current limitations of modern-day approaches \cite{furber2016brain,kendall2020building}, such as deep neural networks trained by the popular backpropagation of errors (or backprop)\cite{rumelhart_learning_1986}. This goal is complementary to (and, to an extent, even a precursor to some elements of) the domain of neurorobotics \cite{kawato1992computational,miyamoto1988feedback}, which focuses on designing robotic devices that contain control systems based on or are inspired by principles of animal/human nervous systems and/or brain structures guided by the key premise that (neural) models are embodied in a body and an environment. While the gap between neurorobotics and many brain-inspired approaches largely is largely divided between focus on real-world hardware (the former) or software simulation (the latter), one pathway to bridging this gap might lie in developing powerful brain-inspired approaches that scale up to and operate robustly on problems that may ultimately be tackled by embodied robotic systems as well as using higher-quality, more realistic simulation platforms (as we do in this work). It is along this path that this work takes a step forward by developing a neurobiologically-grounded neural circuit that is used to craft a complete agent that can tackle \textbf{extremely sparse reward} learning control problems (tested on a more realistic, higher quality robotic system simulator), a problem that many robotic systems must ultimately face, much as humans and animals do in the real world. 

To build such building neural blocks and an agent system, we start from two neurocognitive theoretical foundations, predictive processing (or coding) and planning-as-inference. With respect to predictive coding, which views the brain as a type of hierarchical, pattern-creation engine \cite{friston2006free} that engages in continual self-correction \cite{ororbia2020continual}, we implement a fundamental circuit where each of its levels/regions are implemented by clusters of neurons that attempt to predict the state of other neural clusters/regions and adjust their synapses based on how different their predictions were from observed signals. This allows us to sidestep many of the key issues central to backprop, such as the vanishing/exploding gradient problems \cite{pascanu2013difficulty}, the requirement for a long, unstable credit assignment feedback pathway \cite{ororbia2018biologically}, forward and backward locking problems \cite{jaderberg2016decoupled}, and the need for differentiability \cite{lee2015targetprop,ororbia2018biologically}. 
On the other hand, motivated by planning-as-inference (PAI) \cite{botvinick2012planning} (or active inference (AInf) \cite{friston2011action}), which posits that biological organisms (such as animals and humans) learn a probabilistic generative model by interacting with their world, adjusting the model's internal states to better account for the evidence that the agent acquires over time (unifying perception, action, and learning by framing them as processes that result from approximate Bayesian inference), 
we develop specialized neural circuits that iteratively craft a dynamic generative model of an agent's environment and are further used to produce signals that effectively handle the exploration-exploitation trade-off inherent to reinforcement learning (RL).
Our computational agent framework, \textbf{active predictive coding (ActPC)}, combines both of these concepts/ideas to build a multi-circuit agent that tackles various (simulated) robotic control tasks learning online with extremely sparse feedback.



In light of the above, to summarize, this work makes the following contributions:
\begin{itemize}[noitemsep,nolistsep]
    \item We propose a novel brain-inspired architecture/agent, which we call active predictive coding (ActPC), composed of predictive processing NGC circuits, which conducts a form of online active inference and is trained without using \textbf{backpropagation} of errors. Critically, ActPC balances an epistemic signal (that drives intelligent, non-random exploration) with a goal orienting signal produced by a prior preference circuit (which emulates part of the basal ganglia's role in producing dopamine reward signals), stabilizing the learning process further through a form of action module refresh. 
    \item Our experimental results demonstrate that our ActPC agent, in comparison to several powerful backprop-based RL approaches, performs robustly on a set of robotic control problems, including two difficult realistically-simulated ones taken from the robosuite environment \cite{zhu2020robosuite}. Our results crucially show that ActPC, in general, can learn stably and generalizes well in the face of extremely sparse reward signal problems.
\end{itemize}

\section{Active Predictive Coding}
\label{sec:actpc}


\noindent
\textbf{Notation and Problem Framing:} 
%
In this work, 
$\leftarrow$ is assignment, $\odot$ is a Hadamard product, 
$\cdot$ is a matrix/vector multiplication (or dot product if operators are vectors of the same shape), and $\mathbf{v}^\mathrm{T}$ is the transpose. In the appendix, we have provide a notation table to help readers better understand the components involved in the design of our proposed novel system.
With respect to the problem framing, we consider the standard RL setup for robotic control: an agent interacts with an environment over discrete time-steps -- at $t$, given observation $\mathbf{o}_t \in \mathcal{R}^{D\times1}$, the agents takes action $\mathbf{a}_t$ and then receives reward $r_t$ and observation $\mathbf{o}_{t+1}$. The end goal is to acquire a policy that maximizes expected future reward.

\noindent
\textbf{Problem Setup:} At time $t$, the agent receives an observation $\mathbf{o}_t$ and produces an action (vector) $\mathbf{a}_t$, receiving a sparse scalar reward $r_t$, or a $1$ if it reaches some defined goal state(s) and $0$ or $-1$ otherwise. In all environments 
we study/simulate, the actions are to be real-valued $\mathbf{a}_t \in \mathcal{R}^{A \times 1}$. In general, the environment may be partially 
observed such that all previously generated observation-action pairs $\mathbf{s}_t = (\mathbf{o}_1,\mathbf{a}_1,...,\mathbf{o}_{t-1},\mathbf{a}_{t-1},\mathbf{o}_t)$ 
might be required to fully describe its state. In the simulations investigated in this paper, we assume an identity mapping between been environment 
states and observations, i.e., $\mathbf{s}_t = \mathbf{o}_t$. 
The agent's behavior, defined by policy $\pi$, maps states to a probability distribution over actions $\pi: \mathcal{S} \mapsto p(\mathcal{A})$ (where 
$\mathcal{S}$ is the space of all states and $\mathcal{A} \in \mathcal{R}^{A}$ is the space of all actions). We follow in line with the typical RL approach and model the environment $E$ (which could be stochastic) as a Markov decision process with state space $\mathcal{S}$, action space $\mathcal{A}$, initial state distribution $p(\mathbf{s}_1$, transition dynamics $p(\mathbf{s}_{t+1}|\mathbf{s}_t,\mathbf{a})t)$, as well as reward function $r(\mathbf{s}_t,\mathbf{a}_t)$. Given a state, the return is the sum of discounted future rewards $r_t = \sum^T_{i=1} \gamma^{i-t} r(\mathbf{s}_i,\mathbf{a}_i)$ where $\gamma \in [0,1]$ is the discount factor. Reward values depend on the actions taken and therefore depend upon the agent's policy $\pi$. The general goal is to acquire a policy that maximizes the expected future reward from a start distribution $\mathbb{E}_{r_i,\mathbf{s}_i \sim E, \mathbf{a}_i \sim \pi}[r_1]$.

\subsection{The Neural Generative Coding Base Circuit}
\label{sec:ngc_circuit}

Neural generative coding (NGC) is an instantiation of the predictive processing brain theory \cite{rao1999predictive,friston2005theory,clark2015surfing,ororbia2022neural}, yielding an efficient, robust form of predict-then-correct learning and inference. An NGC circuit in ActPC receives two sensory vectors, an input $\mathbf{x}^i \in \mathcal{R}^{I \times 1}$ ($I$ is the input dimensionality) and an output $\mathbf{x}^o \in \mathcal{R}^{O \times 1}$ ($O$ is the output or target dimensionality). Compactly, an NGC circuit is composed of $L$ layers of feedforward neuronal units, i.e., layer $\ell$ is represented by the state vector $\mathbf{z}^\ell \in \mathcal{R}^{H_\ell \times 1}$ containing $H_\ell$ total units. Given the input--output pair of sensory vectors $\mathbf{x}^i$ and $\mathbf{x}^o$, the circuit clamps the last layer $\mathbf{z}^L$ to the input, i.e, $\mathbf{z}^L =\mathbf{x}^i$, and clamps the first layer $\mathbf{z}^0$ to the output, i.e.,  $\mathbf{z}^0 =\mathbf{x}^o$. Once clamped, the NGC circuit will undergo a settling cycle where it processes the input and output vectors for $K$ steps in time, i.e., it process sensory signals over a stimulus window of $K$ discrete time steps. The activities of the internal neurons (all neurons in between the clamped layers, i.e., $\ell = L-1 \ldots 1$) are updated as follows:
\begin{align}
    \mathbf{z}^\ell \leftarrow  \mathbf{z}^\ell + \beta \Big(-\gamma \mathbf{z}^\ell + (\mathbf{E}^\ell \cdot \mathbf{e}^{\ell-1}) \otimes \partial \phi^\ell(\mathbf{z}^\ell)  -\mathbf{e}^\ell \Big) \label{eqn:state_update}
\end{align}
where $\mathbf{E}^\ell$ is a matrix containing error synapses that pass along mismatch signals from layer $\ell-1$ to $\ell$ (this can be learnable or set to the scaled transpose of the predictive synaptic matrix, i.e., $\mathbf{E}^\ell = \lambda_e (\mathbf{W}^\ell)^T$).
$\beta$ is the neural state update coefficient and set according to $\beta = \frac{1}{\tau}$, where $\tau$ is the integration time constant in the order of milliseconds. This update equation indicates that a vector of neural activity changes, at each step within a settling cycle, according to (from left to right in Equation \ref{eqn:state_update}), a leak term (modulated by the factor $\gamma$), the bottom-up pressure from mismatch signals in lower level neural regions, and a top-down pressure from the neural region above. $\mathbf{e}^\ell \in \mathcal{R}^{H_\ell \times 1}$ are an additional population of special neurons that are tasked entirely with calculating mismatch signals at a layer $\ell$, i.e., $\mathbf{e}^\ell = \mathbf{z}^\ell - \mathbf{\bar{z}}^\ell$, the difference between a layer's current activity and an expectation produced from another layer. Specifically, the layer-wise prediction $\mathbf{\bar{z}}^\ell$ is computed as follows:
\begin{align}
    \mathbf{\bar{z}}^\ell = g^\ell \big( \mathbf{W}^{\ell+1} \cdot \phi^{\ell+1}( \mathbf{z}^{\ell+1} ) + \alpha_m (\mathbf{M}^{\ell+1} \cdot \mathbf{m}_t) \big) \\
    \mathbf{m}_t = \Big[( \mathbf{k}_{t-(H-1)},...,\mathbf{k}_{t-i},...,\mathbf{k}_{t-1} \Big] \mbox{ and }  \mathbf{k}_t = \mathbf{Q} \cdot \mathbf{x}_t
\end{align}
where $\alpha_m = 1$ (but can be set to $0$ if working memory is not desired), $\mathbf{W}^\ell$ denotes the matrix of predictive/generative synapses, $\mathbf{M}^\ell$ is a matrix containing (conditional) memory synapses, and $\mathbf{Q}$ is a random synaptic projection matrix (which each element initialized by a centered Gaussian and standard deviation $\sigma_q$).  $\mathbf{m}_t$ is a working memory vector containing a representation of a recent history of observation, i.e., in this work, $\mathbf{m}_t \in \mathcal{R}^{(I H) \times 1}$ is the concatenation of a small history of randomly projected observations (this paper sets $H = 7$ inspired by classical work in human working memory \cite{miller1956magical}). We found that introducing and generalizing an NGC circuit to operate with a small working memory improved predictive performance and learning stability for inherently time-varying problems such as those encountered in robotics (and in our simulations).
$\phi^{\ell+1}$ is the activation function (and $\partial \phi^{\ell+1}$ is its derivative) for state variables and $g^{\ell}$ (set to be the identity) is applied to predictive outputs. 

\begin{figure*}[!t]
     \centering
     \begin{subfigure}[t]{0.385\textwidth}
        \vskip 0pt
        \centering
         \includegraphics[width=1.05\textwidth]{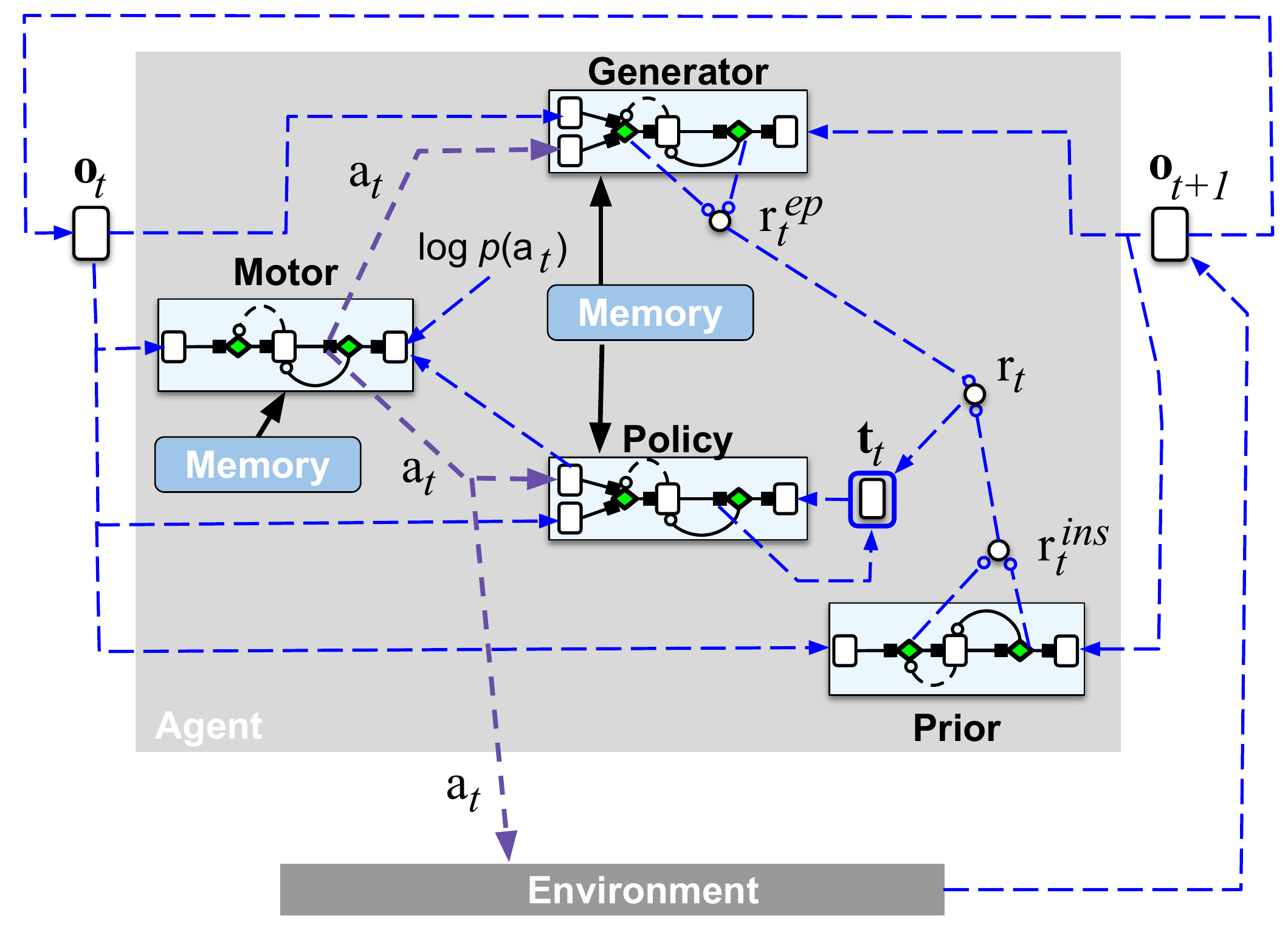}
     \end{subfigure}
     \hspace{0.1cm}
     \begin{subfigure}[t]{0.385\textwidth}
        \vskip 0pt
        \centering
         \includegraphics[width=0.53\textwidth]{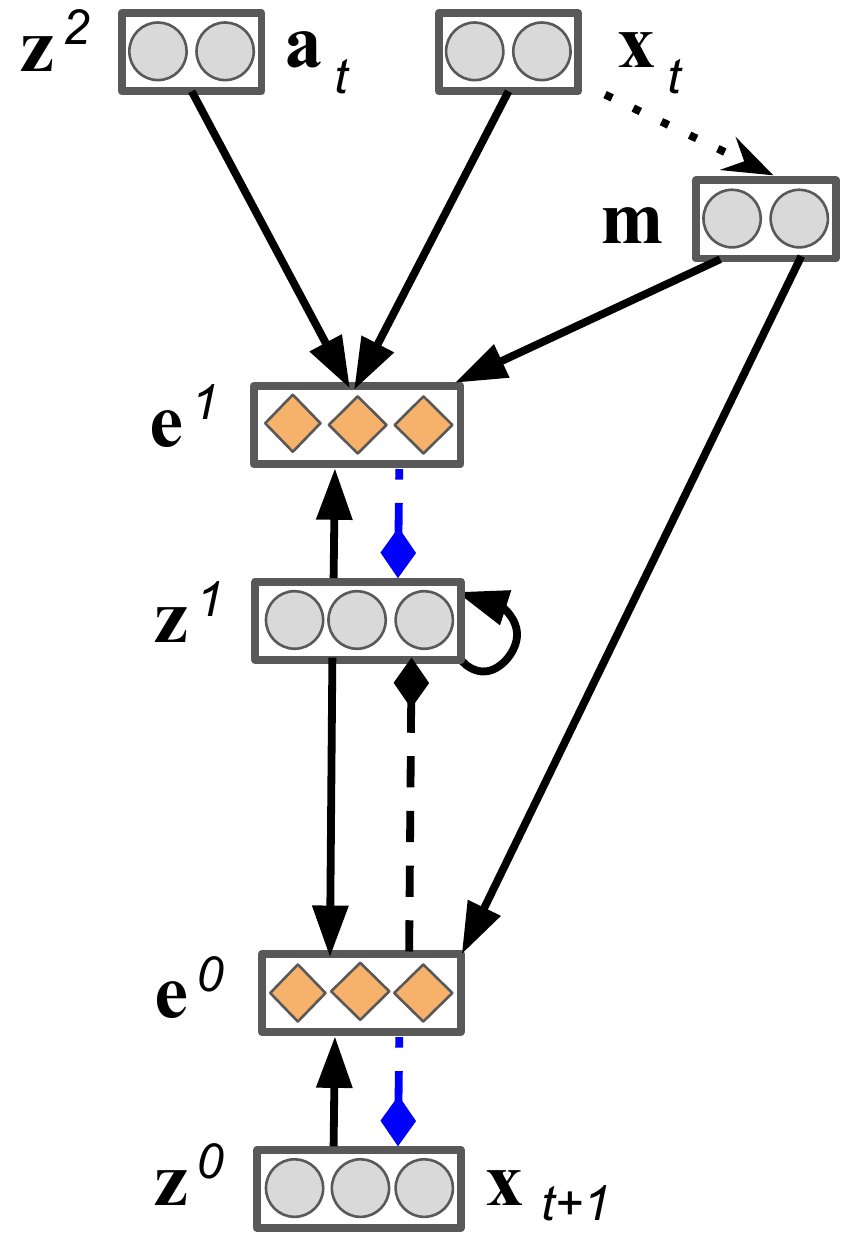}
     \end{subfigure}\\
     \vspace{0.1cm}
     \begin{subfigure}[t]{0.20\textwidth}
        \vskip 0pt
        \centering
         \includegraphics[width=0.8\textwidth]{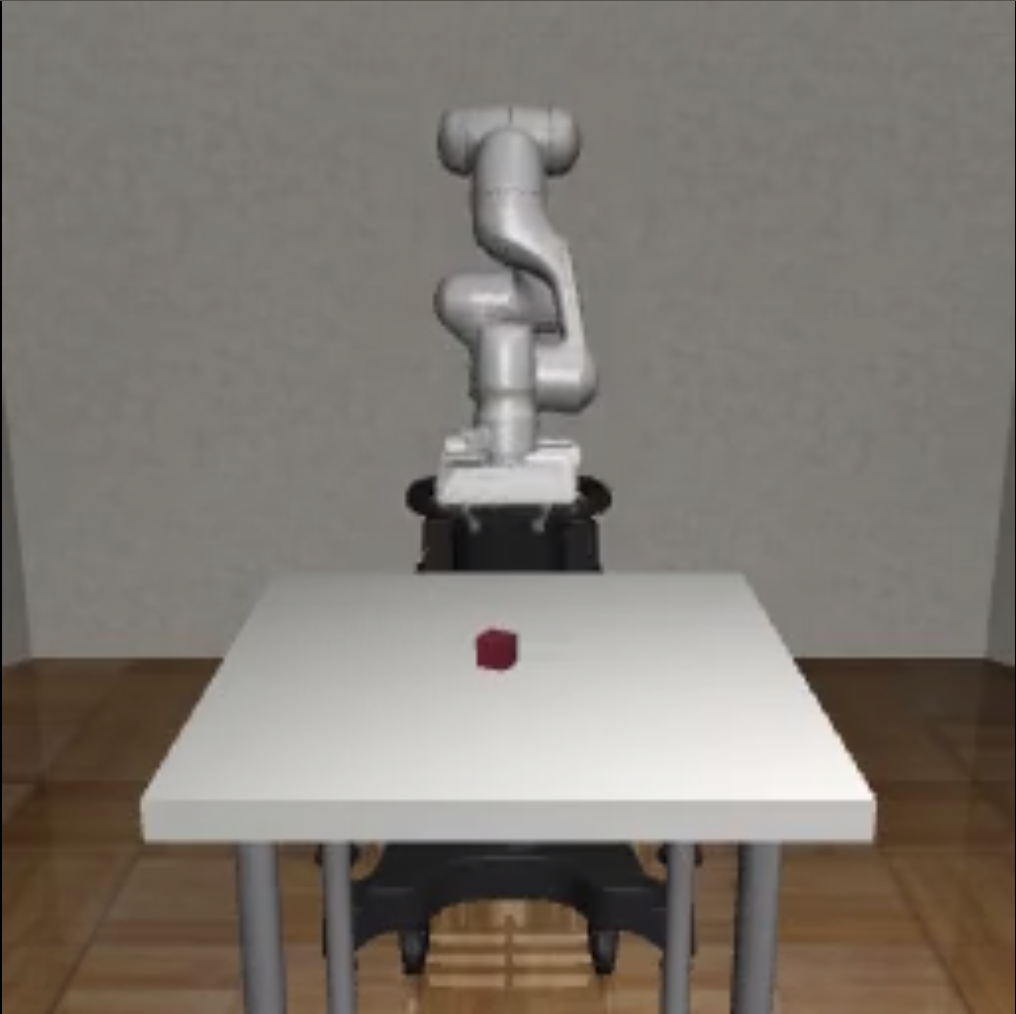}
     \end{subfigure}
     \begin{subfigure}[t]{0.20\textwidth}
        \vskip 0pt
        \centering
         \includegraphics[width=0.8\textwidth]{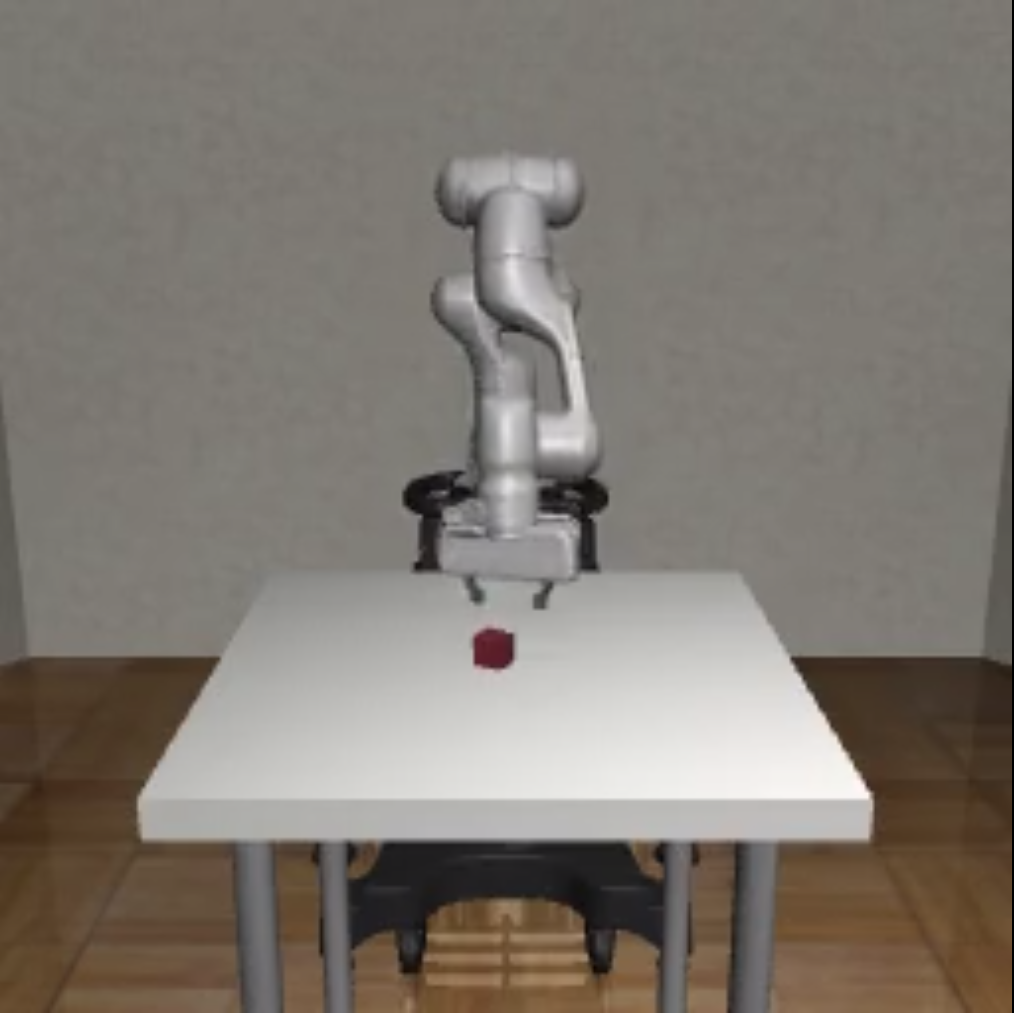}
     \end{subfigure}
     \begin{subfigure}[t]{0.20\textwidth}
        \vskip 0pt
        \centering
         \includegraphics[width=0.8\textwidth]{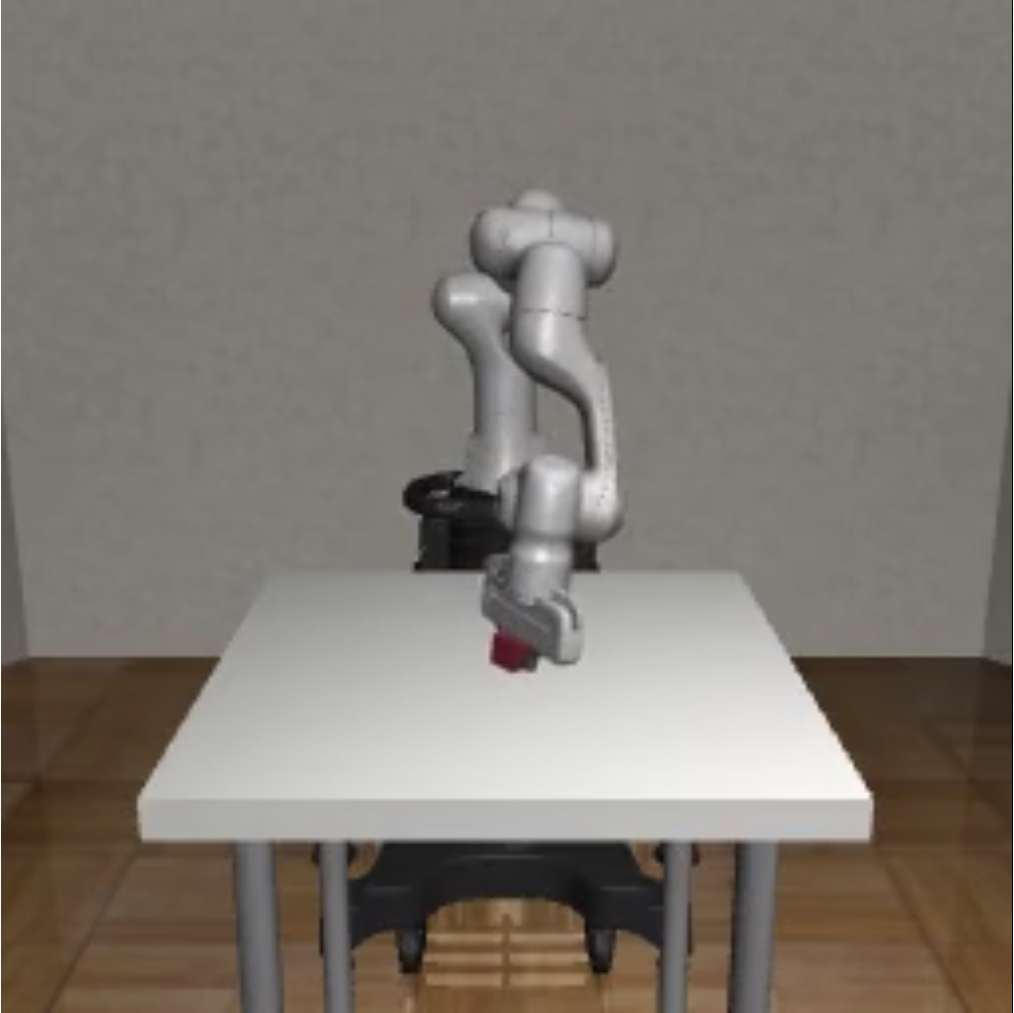}
     \end{subfigure}
     \begin{subfigure}[t]{0.20\textwidth}
        \vskip 0pt
        \centering
         \includegraphics[width=0.8\textwidth]{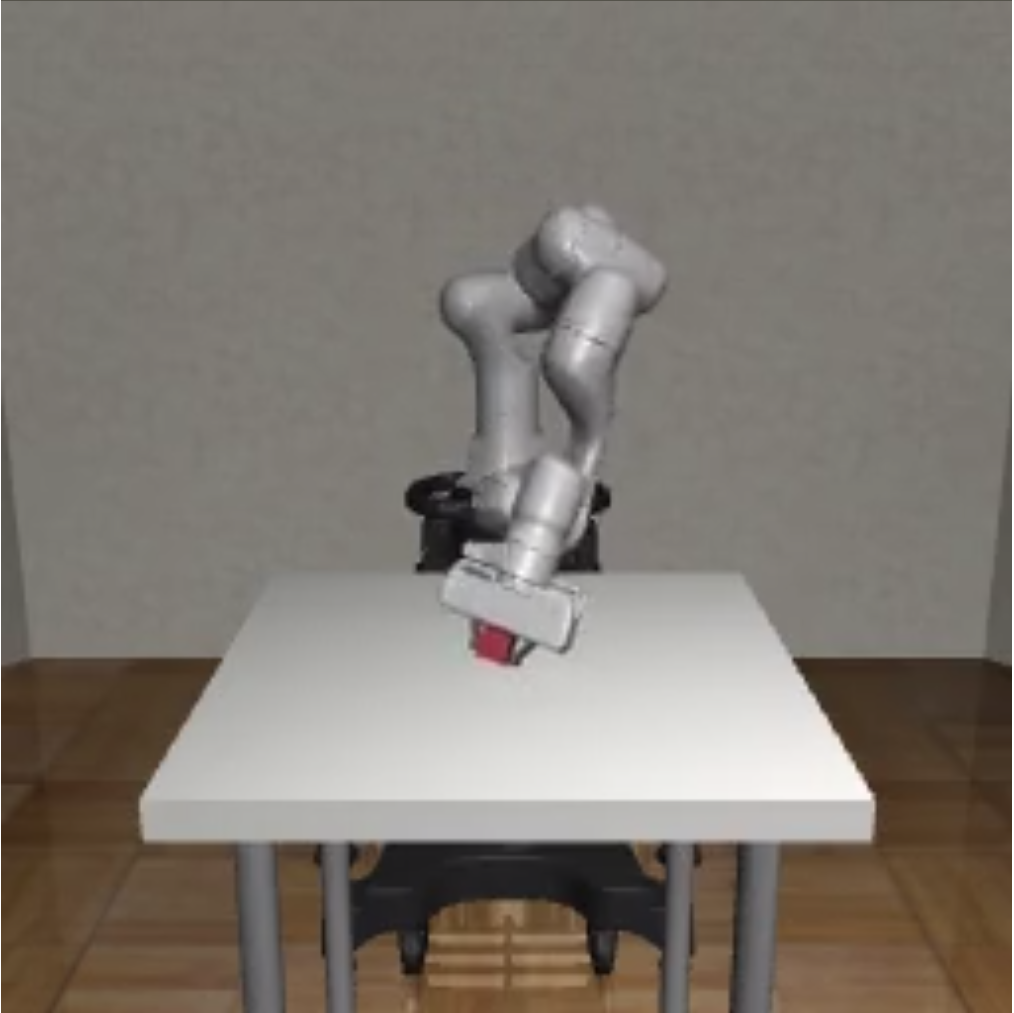}
     \end{subfigure}
\caption{ (Top Left) Depiction of the proposed ActPC architecture. (Top Right) A closer look at one of the ActPC's NGC circuits, i.e., a 3-layer version of our generator/dynamics circuit coupled to a working memory. (Bottom) A trained ANGC robotic system successfully picking up a block (for a robosuite task).}
\label{fig:robo_angc}
\vspace{-0.6cm}
\end{figure*}

After the circuit processes the input--output pair for $K$ steps (repeatedly applying Equation \ref{eqn:state_update} $K$ times), the synaptic matrices are adjusted with a Hebbian-like update rule:
\begin{align}
    \Delta \mathbf{W} &= \mathbf{e}^\ell \cdot (\phi^{\ell+1}( \mathbf{z}^{\ell+1}) )^T \odot \mathbf{M}_W \label{eqn:predictor_update} \\
    \Delta \mathbf{E} &= \gamma_e (\Delta \mathbf{W})^T \odot \mathbf{M}_E \label{eqn:error_update} 
\end{align}
where $\gamma_e$ is a factor ($<1$) to control the time-scale of the error synaptic evolution (ensuring they change more slowly than the predictive ones). $\mathbf{M}_W$ and $\mathbf{M}_E$ are modulation matrices that perform synaptic scaling to ensure additional stability in the learning process \cite{ororbia2022backpropfree}. Note that all circuits in ActPC are implemented according to the mechanistic process described above. We note that the state dynamics and update rules for an NGC circuit can be derived from the objective function that it attempts to optimize, i.e., total discrepancy \cite{ororbia2022neural} which approximates free energy \cite{gershman2019does} (please see Appendix for details). 

Another key function of an NGC circuit is its ability to ancestrally project a vector (akin to a feedforward pass, i.e., no settling process required) through the underlying directed generative model -- we represent this process as $f_{proj}(\mathbf{x}^i; \Theta)$. Formally, ancestrally projecting a vector $\mathbf{x}^i$ proceeds as follows:
\begin{align}
    \mathbf{z}^\ell = \mathbf{\bar{z}}^\ell &= g^\ell( \mathbf{W}^{\ell+1} \cdot \phi^{\ell+1}( \mathbf{z}^{\ell+1} ) ), \; \forall \ell = (L-1),...,0 \label{eqn:projection}
\end{align}
where $\mathbf{z}^L = \mathbf{x}^i$ (only the input/top-most layer is clamped to a specific vector, such as current input $\mathbf{x}^i$). 

\subsection{The NGC Policy Model}
\label{sec:policy_model}
Much as is done in actor-critic/policy approaches, the role of our policy circuit is to describe the expected reward for taking a given action $\mathbf{a}_t$ at observation $\mathbf{o}_t$.
The policy circuit should critique the actions chosen by the agent (or the motor-action circuit, described in the next sub-section), ultimately allowing the agent to produce actions that maximize long-term reward. Inspired by deterministic policy gradients (DPG) \cite{silver2014deterministic} (and its deep backprop-based generalization \cite{lillicrap2015continuous}), which proves useful for optimizing policies over continuous action spaces, we design an NGC circuit that plays the role of ``actor'' (which we label the motor-action model, inspired by the functionality of the human motor cortex) and another circuit that serves as the policy or ``critic''. We introduce further stability as is done in DPG-based approaches by incorporating a target circuit for both the actor and policy (both updated via Polyak averaging).

Our agent's policy circuit induces the following dynamics (within the NGC formulation shown earlier):
\begin{align}
\mathbf{\bar{z}}^2 & = \mathbf{W}^3_{a} \cdot \mathbf{a}_t + \mathbf{W}^3_z \cdot \mathbf{z}^3_t + \mathbf{b}^2 \label{eqn:policy_input} \\ 
\mathbf{\bar{z}}^1 & = \mathbf{W}^2 \cdot \phi( \mathbf{z}^2_t ) + \mathbf{b}^1 \\
\mathbf{q}_{t} = \mathbf{\bar{z}}^0 & = \mathbf{W}^1 \cdot \phi( \mathbf{z}^1_t ) + \mathbf{b}^0 
\end{align}
where we observe that the coefficient $\alpha_m = 0$ (we found that the policy did not need memory $\mathbf{m}_t$ in early experiments and thus omit it to reduce unneeded computation). Note that, for this and all later described circuits, for simplicity of presentation, we present four-layer circuit equations. The parameters of the policy circuit are collected into the construct $\Theta_c = \{\mathbf{W}^1,\mathbf{W}^2,\mathbf{W}^3_{a},\mathbf{W}^3_{z},\mathbf{M}^2,\mathbf{M}^3,\mathbf{b}^0,\mathbf{b}^1,\mathbf{b}^2\}$.

To update the policy model's synaptic efficacies, we then leverage the reward $r_t$ computed by the agent's epistemic/generative and instrumental/prior circuits (described in later sections). Specifically, we compute the target (action-value) vector $\mathbf{t}_t$ by first computing:
\begin{align}
    \mathbf{c}_t &= f_{proj}([\mathbf{a}_t, \mathbf{o}_{t+1}]; \Theta_c) \nonumber \\
    \mathbf{t}_t = \mathbf{z}^0_t &= \begin{cases} 
                    r_t \mathbf{1}_a & \mbox{if } \mathbf{o}_t \mbox{ is terminal} \\
                    r_t + \gamma \mathbf{c}_t &   \mbox{otherwise }
                  \end{cases}          \label{eqn:target_value}
\end{align}
where $\mathbf{1}_a \in 1^{A \times 1}$ (a vector of ones with number of rows equal to the number of different continuous actions).
Once the target vectors have been created, the K-step settling process (Equation \ref{eqn:state_update}) can be executed and all motor-action synapses are updated via Hebbian learning (Equations \ref{eqn:predictor_update}-\ref{eqn:error_update}).

One other special modification was also made to the policy circuit, not apparent in the dynamics equations above -- in contrast to the previous sub-section which introduced one set of error neurons $\mathbf{e}^\ell$ for each layer of the neural circuit, our policy circuit contains two sets (or populations) of error neurons that are triggered depending on whether the policy's or actor's synaptic weights are to updated, i.e., $\mathbf{e}^0 = u_a \mathbf{e}^0_a + (1 - u_a) \mathbf{e}^0_p$, where $u_a = \{0,1\}$ is a binary switch variable to toggle between adjusting policy synapses or actor synapses. 
To update its own synapses ($u_a = 0$), the policy model uses the same error neurons depicted in the Sub-Section \ref{sec:ngc_circuit} (with the target action-value vector. 
On the other hand, when the motor-action circuit's synapses are to be updated ($u_a = 1$), a different set of error neurons are utilized, i.e., $\mathbf{e}^0_a = -\mathbf{1}/A$, which is equivalent to the partial derivative of the average of the policy's actions (if more than one is output) with respect its output nodes. These error neurons allow the policy circuit to be temporarily linked to the actor circuit, permitting it to transmit error messages back to its input action nodes (the first term in Equation \ref{eqn:policy_input}, $\mathbf{W}^3_a \cdot \mathbf{a}_t$) and back on through to the actor circuit. Given that the motor-action's must first traverse through the policy circuit in order to obtain a meaningful output error neuron value, this linkage is necessary (in order to approximately follow the policy gradient) and allows both circuits to run in one single joint settling process. Note that, when the motor-action and policy circuits are linked, the actor's output error neurons are clamped to $\mathbf{e}^0 = \mathbf{d}_a$ where $\mathbf{d}_a = \mathbf{E}^3_a \cdot \mathbf{e}^2$ (or the perturbation that the policy circuit would like to apply to error correct its input action $\mathbf{a}_t$).

\subsection{The NGC Motor-Action Model}
\label{sec:actor_model}

Agents must react as well as manipulate their environment(s). In order to do so, agents need circuits to drive actuators. Building upon the notion of planning-as-inference \cite{botvinick2012planning}, we generalize our NGC circuit to action-driven tasks, particularly to the case of continuous actions (critical/common in the case of robotics). 

Specifically, we design our motor-action model $f_a \colon \mathbf{z}_t \mapsto \mathbf{a}_t$ that outputs a continuous control vector at each time step, i.e., external control signal is $\mathbf{a}_t \in \mathcal{R}^{A \times 1}$ where $A_{ext}$ is the number of different external actions. Action $a_t$ impacts the environment that the ActPC agent is currently interacting with. The motor-action model operates, when given input $\mathbf{z}^3 = \mathbf{o}_t$, formally as follows:
\begin{align}
    \mathbf{\bar{z}}^2 &= \mathbf{W}^3 \cdot \mathbf{z}^3 + \mathbf{M}^3 \cdot \mathbf{m}_t + \mathbf{b}_2 \\
    \mathbf{\bar{z}}^1 &= \mathbf{W}^2 \cdot \phi(\mathbf{z}^2) + \mathbf{M}^2 \cdot \mathbf{m}_t + \mathbf{b}_1 \\
    \mathbf{a}_{t} =  \mathbf{\bar{z}}^0 &= \kappa \text{tanh}(\mathbf{W}^1 \cdot \phi(\mathbf{z}^1) + \mathbf{M}^1 \cdot \mathbf{m}_t + \mathbf{b}_0 )
\end{align}
where $\text{tanh}(\mathbf{v}) = (\exp(2\mathbf{v}) - 1)/(\exp(2\mathbf{v}) + 1)$ and $\kappa$ is a coefficient meant to scale the outputs of the continuous output control signals depending on the lower and upper bounds on the allowable action values for a given problem. 
The parameters of the motor-action circuit are collected into the construct $\Theta_a = \{\mathbf{W}^1,\mathbf{W}^2,\mathbf{W}^3,\mathbf{M}^1,\mathbf{M}^2,\mathbf{M}^3,\mathbf{b}^0,\mathbf{b}^1,\mathbf{b}^2\}$.

To update the synapses for this module, the motor-action circuit must interact with the policy circuit by first running its ancestral projection process (Equation \ref{eqn:projection}) to obtain an immediate action $\mathbf{a}_t$. This produced action is then run through the policy to finally produce (by also ancestrally projecting through the policy neural structure) a set of action-values, one for each continuous action output node. With these action-values produced, the policy's second set of (actor-orienting) output error neurons $\mathbf{e}^0_a$ are used ($\mathbf{e}^0_a = -\mathbf{1}_a/A$) and both the policy and actor circuits are temporarily coupled/linked together and run within one single joint settling process. This joint settling process is done by clamping the motor-action model's output error neurons to the perturbation $\mathbf{d}_a$ produced by the policy circuit (at each step of the settling process) with respect to its input actions, i.e., the actor's output error neurons are set equal to $\mathbf{e}^0 = \mathbf{d}_a$ (ignoring its normally computed $\mathbf{e}^0$). This means that the policy and actor are briefly linked together to produce synaptic updates for the actor circuit (see the Appendix for details and pseudocode). 

Finally, we note that we coupled the motor-action module with its own memory module $\mathcal{M}^{actor}$ based on experience replay \cite{mnih2015human}. However, inspired by work done in self-imitation learning \cite{oh2018self}, we only store in this memory trajectories that reach goal states. Specifically, to store a trajectory into $\mathcal{M}^{actor}$, ActPC measures an episode's cumulative sparse reward, i.e., $r_{epi} = \sum_{t} r_t$, and only stores it if this cumulative signal is greater than/equal to the highest one of the episodes currently stored in $\mathcal{M}^{actor}$ (see Appendix).

\subsection{The Epistemic Signal: The NGC Generative World Model}
\label{sec:dynamics}


Motivated by the finding of expected value estimation in the brain \cite{hikosaka2006basal,rangel2008framework}, ActPC implements a circuit to produce intrinsic reward signals. At a high level, this neural machinery facilitates some  functionality of the basal ganglia and procedural memory, simulating an internal reward-creation process \cite{schultz2016reward}.
Concretely, we refer to the above as an NGC dynamics model, with reward calculated as a function of its error neurons.

In Figure \ref{fig:robo_angc} (Top Right), we graphically depict the design of the NGC dynamics model used to generate epistemic rewards (or intrinsic reward values meant to encourage exploration). 
The NGC dynamics circuit (or generator) processes the current state/observation $\mathbf{o}_t$ and the current action $\mathbf{a}_t$ ($\mathbf{a}_t \in \mathcal{R}^{A \times 1}$ where $A$ is the number of actions), as produced by the motor-action model, and predicts the value of the future state $\mathbf{o}_{t+1}$. When provided with $\mathbf{z}_{t+1}$, the dynamics circuit runs the following for its layer-wise predictions:
\begin{align}
    \mathbf{\bar{z}}^2 & = \mathbf{W}^3_{a} \cdot \mathbf{a}_t + \mathbf{W}^3_z \cdot \mathbf{z}^3_t + \mathbf{M}^3 \cdot \mathbf{m}_t + \mathbf{b}^2 \\ 
    \mathbf{\bar{z}}^1 & = \mathbf{W}^2 \cdot \phi( \mathbf{z}^2_t ) + \mathbf{M}^2 \cdot \mathbf{m}_t + \mathbf{b}^1 \\
    \mathbf{\hat{o}}_{t+1} = \mathbf{\bar{z}}^0 & = g^0\Big(\mathbf{W}^1 \cdot \phi( \mathbf{z}^1_t ) + \mathbf{M}^1 \cdot \mathbf{m}_t + \mathbf{b}^0 \Big)
\end{align}
and leverages the NGC settling process (see Equation \ref{eqn:state_update}) to compute its internal state values, i.e., $\mathbf{z}^3_t, \mathbf{z}^2_t, \mathbf{z}^1_t$. (Note that $\mathbf{W}^3 = [\mathbf{W^3_a, \mathbf{W}^3_z}]$, where we have made clear the explicit distinction between observation conditioned input synapses and action-conditioned input synapses.)
Notice that we have simplified a few items with respect to the NGC circuit -- the topmost layer-wise prediction $\mathbf{\bar{z}}^3_t$ sets $\phi^3(\mathbf{v}) = \mathbf{v}$ for both its top-most inputs $\mathbf{c}^{ext}_t$ and $\mathbf{z}_t$, the post-activation prediction functions for the internal layers are $g^2(\mathbf{v}) = g^1(\mathbf{v}) = \mathbf{v}$, and  $\phi^2(\mathbf{v}) = \phi^1(\mathbf{v}) = \phi(\mathbf{v})$ (the same state activation function type is used in calculating $\mathbf{\hat{z}}^1$ and $\mathbf{\hat{z}}^0$). Once the above dynamics have been executed, the NGC dynamics model's synapses are adjusted via Hebbian synaptic updates. 
The parameters of the generator circuit are collected into the construct $\Theta_g = \{\mathbf{W}^1,\mathbf{W}^2,\mathbf{W}^3,\mathbf{M}^1,\mathbf{M}^2,\mathbf{M}^3,\mathbf{b}^0,\mathbf{b}^1,\mathbf{b}^2\}$.

To generate the value of the epistemic reward \cite{ororbia2022backpropfree}, the dynamics model first settles to a prediction $\mathbf{\hat{z}}_{t+1}$ given the value of the next observation $\mathbf{o}_{t+1}$. After its settling process has finished, the activity signals of its (squared) error neurons are summed to obtain the circuit's  epistemic reward signal:
\begin{align}
    \widehat{r}^{ep}_t &= \sum_j (\mathbf{e}^0 )^2_{j,1} + \sum_j (\mathbf{e}^1 )^2_{j,1} + \sum_j (\mathbf{e}^2 )^2_{j,1} \\
    r^{ep}_t &= \widehat{r}^{ep}_t / (r^{ep}_{max}) \quad \mbox{where } r^{ep}_{max} = \max(\widehat{r}^{ep}_1, \widehat{r}^{ep}_2, ..., \widehat{r}^{ep}_t) \label{eqn:epistemic}
\end{align}
where the epistemic reward signal is normalized to the range of $[0,1]$ by tracking the maximum epistemic signal observed throughout the course of the simulation. 
This epistemic signal encourages the ActPC agent to take actions that explore environmental states/observations that it finds surprising.

\subsection{The NGC Prior:  A Model-Based Instrumental Signal}
\label{sec:model_prior}

To produce a useful instrumental signal to guide the agent to goal states in the face of sparse rewards, without hand-crafted reward functions or human user input, we first learn a generative model of observation states collected from human/machine demonstrations (successful demonstrations on the target problems). This means that the instrumental model can be considered to be a form of imitation learning \cite{hussein2017imitation}.
Formally, our prior model learns to produce $p(\mathbf{o}_{t+1}|\mathbf{o}_t)$ at each step through a given demonstration episode. 

Concretely, our prior circuit elicits the dynamics below:
\begin{align}
    \mathbf{\bar{z}}^2 &= \mathbf{W}^3 \cdot \mathbf{z}^3 + \mathbf{M}^3 \cdot \mathbf{m}_t + \mathbf{b}_2 \\
    \mathbf{\bar{z}}^1 &= \mathbf{W}^2 \cdot \phi(\mathbf{z}^2) + \mathbf{M}^2 \cdot \mathbf{m}_t + \mathbf{b}_1 \\
    \mathbf{\hat{o}}_{t+1} = g^0( \mathbf{\bar{z}}^0 &= \mathbf{W}^1 \cdot \phi(\mathbf{z}^1) + \mathbf{M}^1 \cdot \mathbf{m}_t + \mathbf{b}_0 )
\end{align}
where we see that, in contrast to the epistemic generative circuit, the prior preference module attempts to predict the next observation $\mathbf{o}_{t+1}$ given only the previous observation $\mathbf{o}_t$. 
The parameters of the prior circuit are collected into the construct $\Theta_p = \{\mathbf{W}^1,\mathbf{W}^2,\mathbf{W}^3,\mathbf{M}^1,\mathbf{M}^2,\mathbf{M}^3,\mathbf{b}^0,\mathbf{b}^1,\mathbf{b}^2\}$.

An instrumental reward signal is produced in a similar fashion as that of the epistemic signal:
\begin{align}
    \widehat{r}^{in}_t &= \sum_j (\mathbf{e}^0 )^2_{j,1} + \sum_j (\mathbf{e}^1 )^2_{j,1} + \sum_j (\mathbf{e}^2 )^2_{j,1} \\
    r^{in}_t &= -\widehat{r}^{in}_t / (r^{in}_{max}) \quad \mbox{where } r^{in}_{max} = \max(\widehat{r}^{in}_1, \widehat{r}^{in}_2, ...,\widehat{r}^{in}_t) \label{eqn:instrumental}
\end{align}
where the instrumental reward signal is normalized in the same way as the epistemic signal is as described in the last section. The instrumental signal induces goal-orienting/seeking behavior, mediating the agent's exploration of an environment by ensuring that such exploration serves its intent to solve a given problem or task. Note that the final reward signal $r_t$ is calculated as follows: $r_t = \alpha_{in} r^{in}_t + \alpha_{ep} r^{ep}_t$ where $\alpha_{in}$ and $\alpha_{ep}$ (the reward weighting coefficients) are set to one in this work.



\subsection{Putting It All Together: The Neural Coding Agent}
\label{sec:agent_arch}

Figure \ref{fig:robo_angc} (Top Left) depicts how the various NGC circuits described in the previous sub-sections, e.g., the policy, actor, generator, prior circuits, and the memory replay buffers fit together to compose the agent. 
At a high level, ActPC operates, given observation $\mathbf{o}_t$, according to the following steps:
\textbf{1)} the NGC motor-action circuit takes in $\mathbf{o}_t$ and produces a continuous action $\mathbf{a}_t$,
\textbf{2)} the ActPC agent receives observation $\mathbf{o}_{t+1}$ from the environment, i.e., the result of its action,
\textbf{3)} the NGC generator runs its dynamics ($K$ steps of Equation \ref{eqn:state_update}) to find a set of hidden neural activity values that allow a mapping from $[\mathbf{a},\mathbf{o}_t]$ to $\mathbf{o}_{t+1}$, produces epistemic signal $r^{ep}_t$ (Equation \ref{eqn:epistemic}), and then updates its own specific synapses using Equations \ref{eqn:predictor_update}-\ref{eqn:error_update}, 
\textbf{4)} the NGC prior circuit takes in $\mathbf{o}_t$ and runs its dynamics ($K$ steps of Equation \ref{eqn:state_update}) to produce an instrumental signal $r^{in}_t$ (Equation \ref{eqn:instrumental}), and then the full reward signal is computed as $r_t = \alpha_{in} r^{in}_t + \alpha_{ep} r^{ep}_t$,
\textbf{5)} the NGC policy circuit takes in $\mathbf{a}_t$ and $\mathbf{o}_t$ and runs its dynamics ($K$ steps of Equation \ref{eqn:state_update}) to find a 
set of hidden neural activities that allow a mapping from $\mathbf{o}_t$ to $\mathbf{t}_t$ (which contains $r_t$), and then the actor and policy circuits update their synapses (via Equations \ref{eqn:predictor_update}-\ref{eqn:error_update}), and, finally, 
\textbf{6)} the ANGC agent transitions to $\mathbf{o}_{t+1}$ and moves back to step 1. We refer the reader to the appendix for pseudocode depicting the online mechanics of ActPC, a discussion on connections to more general active inference biological process theory, policy gradients, as well as additional neurobiological connections/motivations behind the agent's design.

In addition to the target circuit modification mentioned earlier, we leveraged combined experience replay \cite{zhang2017deeper} (see Appendix for details), which fits our agent's online processing scheme, where the replay buffers \cite{o2010play,mnih2015human} were implemented as ring buffers with mini-batches sampled uniformly. 

\section{Experimental Results}
\label{sec:experiments}

In this section, we describe some basic details of our experimental setup while the Appendix contains extensive detail related to hyper-parameters, task environment specifications, and related work.

\subsection{Investigated Models}

For the Mujoco robotic control tasks, we explored the following baselines: twin-delayed DDPG (TD3) \cite{td3_18}, hindsight experience replay (HER) \cite{andrychowicz2017hindsight}, and Learning Online with Guidance Offline (LOGO) \cite{logo21} (all models used demonstration data -- see Appendix for descriptions of each as well as specific modifications made for this work). For the robosuite tasks, we compare to DDPG-Demo, which generalizes DDPG \cite{lillicrap2015continuous} to utilize demonstration data (we found it failed to perform without demonstrations).

\subsection{Simulated Tasks}
\label{sec:tasks}


\paragraph{Mujoco Tasks}
We start by experimenting with four challenging Mujoco-simulation tasks, namely the ``reacher'' (Reacher-v4), ``half-cheetah'' (Half-Cheetah-v2), ``hopper'' (Hopper-v2), and ``walker'' (Walker-v2), and ``swimmer'' (Swimmer-v2) tasks. All five tasks are converted to sparse reward variants. For all tasks except reacher, a reward $r_t$ of $1$ is awarded if a particular condition is met during an episode whereas for reacher, $r_t = 1.0$ is awarded to the agent when the L2 distance between the arm fingertip and the goal state is less than $0.06$ (else $r_t = 0$ or $r_t = -1$).

\paragraph{The Robosuite Tasks:}
We next experiment with two difficult tasks taken from the high-quality, realistic simulator known as robosuite. In particular, we simulate the ``block lifting'' (Block Lift) and ``can pick-and-place'' (Can Place) tasks. Each problem is set to their sparse equivalent -- $r_t = 1$ if the robotic arm reaches the goal state, otherwise $r_t = 0$.

\subsection{Relative Stability (R-Stability)}
\label{sec:rstability}
In this work, we also introduce a new and simple metric for characterizing the learning stability (and indirectly the quality of an agent's convergence to a policy in the online learning setting) which we called \emph{relative stability} (r-stability for short, or RS as in below). R-stability is formally measured over a window $\mathcal{W}$ of the (sparse) reward values the last $K_e$ episodes of agent's simulation which we designed to be calculated as follows:
\begin{align}
    \text{RS} = \frac{1}{K_e} \sum^{K_e}_i \bigg|\frac{\mathcal{W}^\mu_i - \max(\mathcal{W})}{\max(\mathcal{W})}\bigg|
\end{align}
where $\mathcal{W}^\mu$ is a window containing the last $K_e$ values the of the smoothened average of $\mathcal{W}$ (each value at index $i$ in $\mathcal{W}^\mu$ is the rolling average of the previous $100$ values before it). In essence, this metric is interpreted as the mean value of the absolute relative errors produced by the agent with respect to the locally maximum/best episodic return (found within the final $K_e$ episodes). In this work, we set $K_e = 100$.
Lower values closer to zero imply better stability, with zero yielding perfect stability. The motivation behind this metric is that agents that converge (or nearly converge) to a useful policy that results in high episodic return should remain near this policy (without major fluctuations or possible collapse towards failure), indirectly how well the agent is able to retain the policy acquired in its memory. 
Note that poorly-performing agents can also obtain a good r-stability score as well (as an agent that hovers around a random or sub-optimal policy would exhibit a lower r-stability measure) -- we recommend measuring this metric in tandem with the agent's average episodic return over the last $K_e$ episodes to ensure that the agent achieves high return and low r-stability (as we do in the main paper).

\begin{figure*}[!t]
     \centering
     \begin{subfigure}[t]{0.385\textwidth}
        \vskip 0pt
        \centering
         \includegraphics[width=1.0\textwidth]{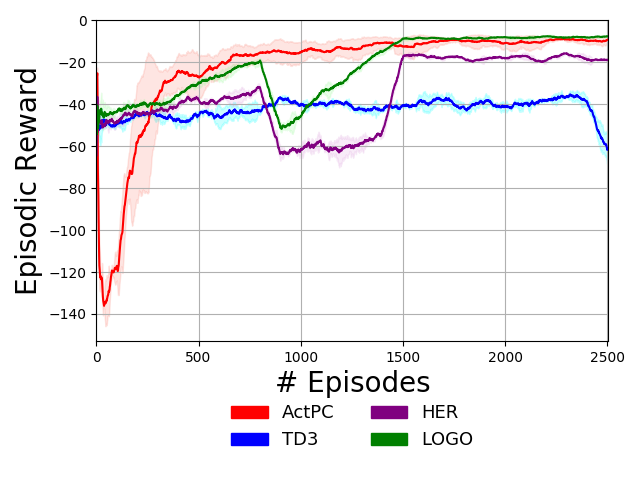}
     \end{subfigure}
     \hspace{0.1cm}
     \begin{subfigure}[t]{0.385\textwidth}
        \vskip 0pt
        \centering
         \includegraphics[width=1.0\textwidth]{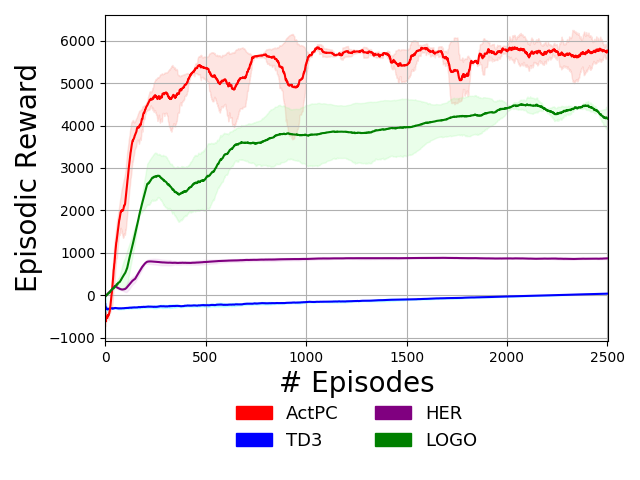}
     \end{subfigure}\\
     \begin{subfigure}[t]{0.31\textwidth}
        \vskip 0pt
        \centering
         \includegraphics[width=1.0\textwidth]{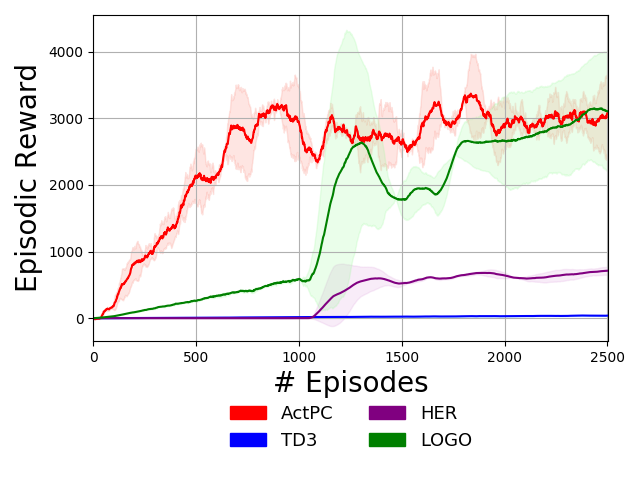}
     \end{subfigure}
     \begin{subfigure}[t]{0.31\textwidth}
        \vskip 0pt
        \centering
         \includegraphics[width=1.0\textwidth]{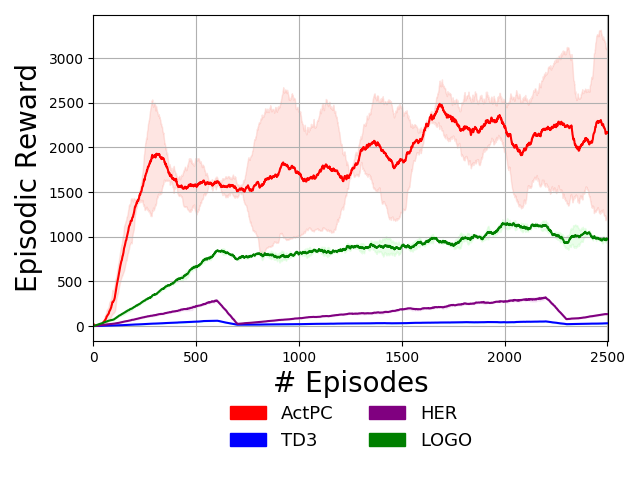}
     \end{subfigure}
     \begin{subfigure}[t]{0.31\textwidth}
        \vskip 0pt
        \centering
         \includegraphics[width=1.0\textwidth]{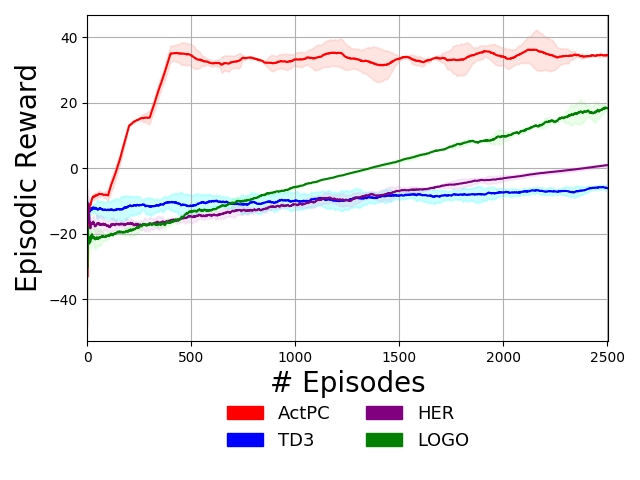}
     \end{subfigure}
\caption{Sparse robotic control problem results: (Top Left) reacher results, (Top Right) half-cheetah results, (Bottom Left) walker, (Bottom Middle) hopper, and (Bottom Right) swimmer (five-trial reward mean and standard deviation).}
\label{fig:mujoco_results}
\vspace{-0.5cm}
\end{figure*}

\begin{figure*}[!t]
     \centering
     \begin{subfigure}[t]{0.385\textwidth}
        \vskip 0pt
        \centering
         \includegraphics[width=1.0\textwidth]{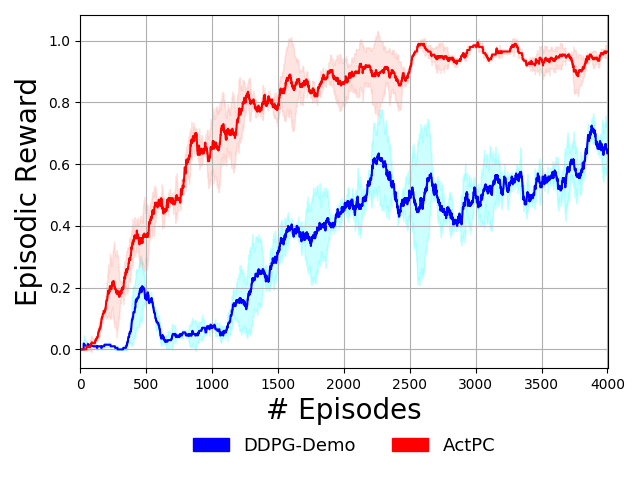}
     \end{subfigure}
     \hspace{0.1cm}
     \begin{subfigure}[t]{0.385\textwidth}
        \vskip 0pt
        \centering
         \includegraphics[width=1.0\textwidth]{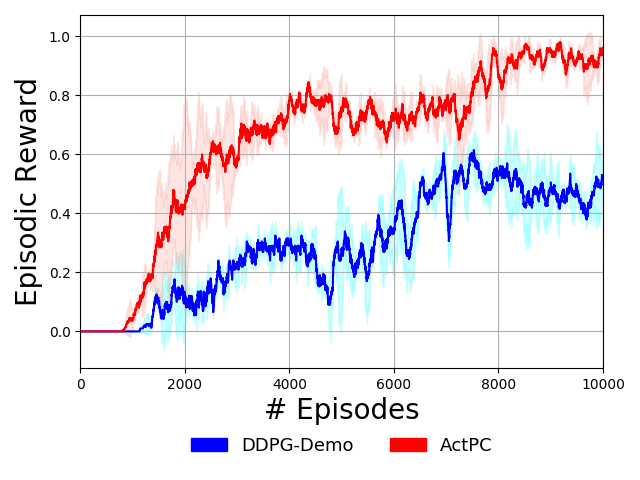}
     \end{subfigure}
\caption{ (Left) The block-lifting task and (Right) can pick-and-place task results (five-trial reward mean and standard deviation reported).}
\label{fig:robosuite_results}
\vspace{-0.5cm}
\end{figure*}

\begin{table}[!t]
\begin{center}
\caption{Mujoco performance results (5-trial mean and standard deviation reported).}
\label{tab:mujoco_metrics}
\begin{tabular}{ |c|c|c| } 
 \hline
 \textbf{Reacher} & \textbf{Avg. Return} & \textbf{R-Stability} \\ 
  \hline
TD3 & $-61.682 \pm 7.227$ & $2.085 \pm 0.323$ \\
HER & $-18.772 \pm 0.578$ & $15.619 \pm 0.357$ \\
LOGO & $-7.583 \pm 0.207$ & $14.956 \pm 1.542$ \\
ActPC & $-9.269 \pm 1.442$ & $3.347 \pm 1.116$ \\
 \hline
  \textbf{Half-Cheetah} & \textbf{Avg. Return} & \textbf{R-Stability} \\ 
  \hline
TD3 & $36.988 \pm 3.893$ & $0.607 \pm 0.026$ \\
HER & $868.9 \pm 19.827$ & $0.046 \pm 0.011$ \\
LOGO & $4163.38 \pm 260.745$ & $0.085 \pm 0.066$ \\
ActPC & $5757.505 \pm 130.922$ & $0.088 \pm 0.036$ \\
 \hline
  \textbf{Walker} & \textbf{Avg. Return} & \textbf{R-Stability} \\ 
  \hline
TD3 & $39.071 \pm 0.025$ & $0.517 \pm 0.022$ \\
HER & $712.948 \pm 57.917$ & $0.119 \pm 0.011$ \\
LOGO & $3104.631 \pm 905.941$ & $0.096 \pm 0.061$ \\
ActPC & $3081.107 \pm 668.267$ & $0.568 \pm 0.02$ \\
 \hline
  \textbf{Hopper} & \textbf{Avg. Return} & \textbf{R-Stability} \\ 
  \hline
TD3 & $30.935 \pm 0.65$ & $0.503 \pm 0.043$ \\
HER & $133.638 \pm 9.655$ & $0.453 \pm 0.037$ \\
LOGO & $975.368 \pm 24.273$ & $0.461 \pm 0.007$ \\
ActPC & $2170.714 \pm 918.793$ & $0.349 \pm 0.227$ \\
 \hline
  \textbf{Swimmer} & \textbf{Avg. Return} & \textbf{R-Stability} \\ 
  \hline
TD3 & $-6.052 \pm 0.678$ & $10.349 \pm 1.81$ \\
HER & $0.956 \pm 0.51$ & $0.779 \pm 0.129$ \\
LOGO & $18.342 \pm 1.221$ & $0.516 \pm 0.036$ \\
ActPC & $34.498 \pm 0.598$ & $0.21 \pm 0.008$ \\
 \hline
\end{tabular}
\end{center}
\vspace{-0.5cm}
\end{table}

\begin{table}[!t]
\begin{center}
\caption{Robosuite performance results (5-trial mean and standard deviation reported).}
\label{tab:robosuite_metrics}
\begin{tabular}{ |c|c|c| } 
 \hline
 \textbf{Block Lift} & \textbf{Acc} & \textbf{S-Dist} \\ 
  \hline 
BC & $100.0 \pm 0.0$ & $--$ \\
BC-RNN & $100.0 \pm 0.0$ & $--$ \\
BCQ & $100.0 \pm 0.0$ & $--$ \\
CQL & $56.7 \pm 40.3$ & $--$ \\
HBC & $100.0 \pm 0.0$ & $--$ \\
IRIS & $100.0 \pm 0.0$ & $--$ \\
DDPG-Demo & $63.5 \pm 7.8$ & $0.340 \pm 0.043$ \\
ActPC & $96.5 \pm 2.1$ & $0.048 \pm 0.008$ \\
 \hline
\textbf{Can Place} & \textbf{Acc} & \textbf{S-Dist} \\ 
\hline
BC & $86.0 \pm 4.3$ & $--$ \\
BC-RNN & $100.0 \pm 0.0$ & $--$ \\
BCQ & $62.7 \pm 8.2$ & $--$ \\
CQL & $22.0 \pm 5.7$ & $--$ \\
HBC & $91.3 \pm 2.5$ & $--$ \\
IRIS & $92.7 \pm 0.9$ & $--$ \\
DDPG-Demo & $51.5 \pm 3.5$ & $0.351 \pm 0.079$ \\
ActPC & $94.0 \pm 2.1$ & $0.101 \pm 0.028$ \\ 
 \hline
\end{tabular}
\end{center}
\vspace{-0.5cm}
\end{table}

\section{Discussion}
\label{sec:discsusion}

In Figures \ref{fig:mujoco_results} (Mujoco) and \ref{fig:robosuite_results} (Robosuite), we plot the smoothened (over a window of $100$ episodes) the average episodic reward as a function of episode. In Tables  \ref{tab:mujoco_metrics} (Mujoco) and \ref{tab:robosuite_metrics}, we report the final ($100$-episode) average reward (Mujoco, Avg. Return) or the average $100$-episode success rate (Acc, Robosuite) in the first column and we measure the relative stability (R-Stability) of each approach (for both tasks) in the second column. As described earlier, R-Stability is our proposed metric for measuring how stable an RL algorithm is towards the end of simulation (lower values closer to zero imply better stability, with zero yielding perfect stability -- see Appendix for formulation). 

Based on the results of our simulations, we find that: (1) ActPC is able to perform well on both sets of tasks for all problems (both Mujoco and robosuite), (2) ActPC outperforms HER and TD3 on all five Mujoco tasks ( and LOGO in 4/5 cases) while outperforming DDPG-Demo on both robosuite tasks (Figures \ref{fig:mujoco_results} \& \ref{fig:robosuite_results}), (3) offers generally good stability for most Mujoco tasks (it is best in 3/5 tasks, if we disregard the fact that TD3 performed poorly on the reacher task with a low return) and best stability for both robosuite tasks, and (4) is competitive with top-performing powerful baselines on the robosuite tasks, nearly reaching the same performance as behavior cloning (BC) and recurrent BC (BC-RNN) \cite{mandlekar2021matters}, outperforming other models such as Conservative Q-Learning (CQL) \cite{mandlekar2021matters} on both tasks, and outperforming models such as hierarchical behavior cloning (HBC) \cite{mandlekar2021matters}, batch-constrained Q-learning (BCQ) \cite{mandlekar2021matters}, and IRIS on the harder can pick-and-place task. 

With respect to the three key baseline models that we experimented with, e.g., TD3, HER, and LOGO, we wanted to highlight that TD3 and HER do not learn in a sparse environment when we do not provide demonstration (demo) data and attempt to train these systems from scratch (without any demonstrations). On the other hand, without demo samples, LOGO takes many episodes to start learning; for instance, in preliminary experimentation, we observed that on the swimmer task that LOGO required a minimum of 3.5k episodes to just get returns/values greater than zero. Hence, to bolster/improve the baselines, we trained a PPO agent on dense rewards for $2000$ episodes to collect demonstration sample trajectories. We ensured that all baseline models had access to the same demo data. 

It is important to note that models such as TD3 and HER struggle to reach optimal performance. Even though HER is designed to work in sparse environments, we found that it struggles in environments where goal states are undefined (such as in the sparse Mujoco problems we explored). According to our results, one of the baseline approaches that comes closer to ActPC is LOGO, which also comes with a theoretical guarantee when using optimal demo data. Nevertheless, we point out that there are a few apparent differences between the processes adapted by LOGO and our implementation of LOGO. First, we used fixed episodes to train the demo data model; however, in the original paper the dense model was tuned more carefully based on the reward signals. As a result, our implementation can be viewed as LOGO with only partially available demo data. Furthermore, in problems such as reacher, our sparse environment setup is slightly different from the ones examined in the original LOGO work given that they provided a positive reward of $+1$ when the agent moved $2$, $20$, and $2$ units from their initial position, respectively. However, in tasks such as reacher, we provide a reward of $+1$ based on the Euclidean distance from the goal state for problems. 

It is worth noting that ActPC, a neuro-cognitive inspired architecture trained without backprop stably works with sparse rewards and is competitive across the  variety of difficult control task environments we examined. As noted throughout the paper, ActPC does not involve any reward manipulation/design or additional environmental signals to ensure stability/learning. The model, by design, ensures relatively prediction stays consistent throughout learning -- this is, again, evident based on $R$-stability analysis as shown in Tables \ref{tab:mujoco_metrics} and \ref{tab:robosuite_metrics} (our agent obtains higher reward with generally lower R-stability measurements).

\noindent
\textbf{Limitations}
Despite the promise that ActPC offers for robotic control (and potentially neurorobotics), the current implementation does have several drawbacks that should be considered in future research. 
To start, if one does not design an NGC circuit with capacity in mind at the beginning, i.e., considering the total number of weights initialized when choosing its number of layers and the number of neural units in each layer, it is very easy to create agents that require greater memory storage. An NGC circuit operates with both forward generative/prediction synapses \textbf{as well as} error correction/feedback synapses, meaning that adding additional processing elements to any single layer increases in required memory for storing the additional forward and error synapses (i.e., larger synaptic parameter matrices). Aside from taking care to design NGC circuits with this increased memory usage in mind, one could set the error feedback synapses to be equal to the transpose of the predictive/forward synapses (giving up the biological plausibility of our asymmetric forward and backward pathways) in order to significantly reduce the memory footprint (although we observe that often more steps of iterative inference, i.e., a higher $K$, is required in the case of tied predictive/error synapses).

Beyond the above, although we have experimentally shown that ActPC can acquire good-quality policies reasonably early when learning dynamically/online compared to several backprop-based RL models, the internal NGC predictive processing circuits do require, per time step (per input observation), additional computation in order to (iteratively) settle to latent state activities that minimize each circuit's total discrepancy. Fortunately, for the problems studied in this work, we found that the ActPC's NGC circuits required a small number of steps between $K=10$ and $K=25$), for more complex patterns of a higher dimensionalities, it is likely that a higher value of $K$ will be required in order to settle to useful latent states (before triggering each circuit's local Hebbian synaptic updates). 
Beyond designing software/hardware that exploits the natural layer-wise parallelism afforded by NGC's inference and learning computations (given that NGC is not forward or update-locked each layer of neurons/synapses could be placed on their own set of GPUs/CPUs), we note that an important direction for combatting potential growing costs for an NGC circuit's iterative settling process is utilize amortized inference (as is done in the case of predictive sparse decomposition \cite{kavukcuoglu2010fast}), which would greatly reduce the number of settling steps required by learning a fast initialization information propagation model alongside the normal circuit synapses (at the cost of requiring additional memory to store the initialization model parameters). 

Finally, it still remains to be seen how a generalization of ActPC towards partially observable Markov decision process (POMDPs) would work. How does the ActPC system operate when two additional circuits, specifically an encoder and decoder (which would remove this work's current assumption that the mapping between observations and latent states is the identity matrix), are introduced? This would also warrant investigation into how NGC circuits function, such as when dealing with raw pixels/natural images, when using locally-connected weight structures and convolution operators instead of fully-connected, dense neural structures. 
Note that work in this direction \cite{ororbia2020large} offers some potentially useful local update rules for convolution but consideration on how to efficiently integrate convolution will be key and a methodology for quickly deciding a reasonable number of filters for acquiring useful feature detectors will be needed (much as is done with modern-day convolutional neural networks).

\section{Conclusions and Future Work}
\label{sec:conclusion}

In this paper, we proposed active predictive coding (ActPC), a neurobiologically-inspired learning process and agent system for tackling challenging sparse reward robotic control problems. Our experimental results, across two sets of robotic tasks 
demonstrate that it is indeed possible to learn powerful, goal-directed control agents without backpropagation of errors (backprop). Furthermore, our work provides further evidence that a neuro-biologically grounded scheme can offer a promising pathway towards developing  agents that tackle the exploration-exploitation tradeoff dynamically, which could prove useful for research domains such as neurorobotics. ActPC, while potentially powerful as our results indicate, is not without its limitations including other biological elements that are still missing, such as neural communication through spike trains (although related work has demonstrated that neural generative coding can be reformulated as a spiking neural system \cite{ororbia2019spiking}, a direction we intend to explore in future work).
Future directions will include further studying and expanding the types of robotic control tasks ActPC can tackle, investigating how ActPC can tackle the partially observable Markov decision process formulation of robotic control environments, and crucially generalizing ActPC to deal with the domain adaptation problem inherent to creating an embodied robotic agent for tackling real-world tasks.


\bibliographystyle{IEEEtran}
\bibliography{ref}

\section{Appendix}
\section*{Simulated Task Specifications}
\label{sec:task_details}

In this section, we provide details for the Mujoco and robosuite control tasks that we simulated in the main paper.

\subsection*{Mujoco Robotic Control Tasks}

\noindent
\textbf{Reacher:} The state space for this environment is $11$-dimensional and includes the position and velocity of each joint. The initial states are uniformly randomized. The action space is a two-dimensional continuous space. A sparse reward is provided in which reward of $1.0$ is awarded to the agent when the (Euclidean) distance between the arm fingertip and the target/goal state is less than $0.06$, otherwise it is either $0$ or $-1$ (baselines used $-1$ given that we observed improved performance when using this). 

\noindent
\textbf{Half-cheetah:} The state space is $17$-dimensional and includes the position and velocity of each joint and the initial states are uniformly
randomized. The action space is continuous and 6-dimensional. A sparse reward is provided where a reward of $1.0$ is given whenever the agent meets certain conditions that make the two-dimensional (2D) cheetah system run forward (right) as fast as possible (the sparsified version of this problem makes this signal quite sparse) and zero otherwise.

\noindent
\textbf{Walker:} The state space is 17-dimension, position and velocity of each joint. The initial states are uniformly randomized.
The action is a 6-dimensional continuous space. A sparse reward is used in which reward of $1.0$ is awarded to the agent whenever it moves the 2D, two-legged figure forward (zero otherwise) -- to do so, the agent must coordinate both sets of feet, legs, and thighs to move forward (to the right).

\noindent
\textbf{Hopper:} The state space is 11-dimension, position and velocity of each joint. The initial states are uniformly randomized.
The action is a 3-dimensional continuous space. The agent's goal is to make the 2D, one-legged robot hop such that it moves forward (right) by applying torques on the three hinges connecting the system's four body parts (a reward of  $1.0$ is given if the system moves forward, zero otherwise). 

\noindent
\textbf{Swimmer:} The state space is 8-dimension, position and velocity of each joint. The initial states are uniformly randomized.
The action is a 2-dimensional continuous space. The agent's goal is to swim the 2D figure robot, which is suspended in a two-dimensional ``water'' pool, towards the right by applying torque on the rotors and leveraging the friction of the fluid (a reward of  $1.0$ is given if the system moves to the right, zero otherwise).

\subsection*{Robosuite Robotic Control Tasks}

For both robosuite tasks, the robotic model we used was the 7 degrees-of-freedom Panda arm. It was set to utilize a parallel-jaw gripper containing two smaller finger pads.

\noindent
\textbf{Block Lifting:} In this task, a robotic arm must pick up a cube placed on the tabletop in front of it and lift above a certain height. The cube location is randomized at the start of each and every episode. The state-space contains $42$ dimensions and the action-space is of size $7$. A sparse reward of $1$ is provided whenever the agent succeeds to pick up and lift the cube correctly and $0$ otherwise.

\noindent
\textbf{Can Pick-and-Place:} A (beer) can is placed into a bin in front of the robotic arm and the agent must pick up and place the can object into its corresponding container next to the bin. The can location is randomly changed at the beginning of every episode. 
The state-space contains $46$ dimensions and the action-space is of size $7$. A sparse reward of $1$ is provided whenever the agent succeeds in placing the can into its correct bin and $0$ otherwise.

\subsection*{Demonstration Data}

For both the Mujoco and robosuite tasks, we crafted small buffers containing demonstration samples and each baseline and the ActPC utilized these samples. For the Mujoco tasks, the baselines had access to much larger demonstration data pools ($2000$ successful trajectories, collected from a trained PPO model using a shaped dense reward) while ActPC only had access to $300$ samples (also collected from a trained PPO that used a dense shaped reward) to learn its prior preference and sample from to lightly guide the motor-action model. For the robosuite tasks, we used the (multi-modal) multi-human (MH) dataset collected from the robomimic project \cite{mandlekar2021matters} and both the DDPG-Demo and ActPC had access to the same $300$ human trajectories (of varying proficiencies/abilities).

\section*{Neural Generative Coding Dynamics}


\noindent
\textbf{Definition/Notation Table:} In Table \ref{table:definitions}, we explain what each mathematical acronym/symbol/operation/abbreviation in the main paper represents.

\begin{table}[!t]
\caption{Table of key symbol/operator/abbreviation definitions.}
\label{table:definitions}
\begin{center}
\begin{tabular}{||c c||} 
 \hline
 \textbf{Item} & \textbf{Explanation} \\ [0.5ex] 
 \hline\hline
 ActPC & Active predictive coding (model) \\ 
 \hline
 NGC & Neural generative coding (circuit model) \\
 \hline
 $\mathbf{v} \in \mathcal{R}^{D \times 1}$ & A column vector $\mathbf{v}$ of shape $D \times 1$ \\
 \hline
 $\mathbf{M} \in \mathcal{R}^{B \times D}$ & A matrix $\mathbf{M}$ of shape $B \times D$ \\
 \hline
 $\cdot$ & Matrix/vector multiplication \\
 \hline
 $\odot$ & Hadamard product (element-wise multiplication) \\
 \hline
 $(\mathbf{v})^T$ & Transpose of $\mathbf{v}$ \\
 \hline
 $||\mathbf{v}||_2$ & Euclidean norm of $\mathbf{v}$ \\
 \hline
 $E$ & The environment to simulate (e.g., a robotic control task)\\
 \hline 
 $\mathbf{o}_t$ & The observation of environment $E$ at time step $t$ \\
 \hline
 $\mathbf{r}_t$ & The scalar (1x1) reward signal return by environment $E$ at time $t$ \\
 \hline
 $\mathbf{r}^{in}_t$ & The instrumental, or goal-orienting, signal (ActPC wants to maximize $-\mathbf{r}^{in}_t$ - find goal-path states) \\
 \hline
 $\mathbf{r}^{ep}_t$ & The epistemic, or exploration-inducing, signal (ActPC wants to maximize $-\mathbf{r}^{ep}_t$ - find surprising states) \\
 \hline
 $H_\ell$ & The number of neurons in the $\ell$-th layer of the NGC generative circuit. \\
 \hline
 $N_e$ & The number of episodes an agent is simulated on (w/ environment $E$)\\
 \hline
 $T$ & The maximum number of discrete time-steps per simulated episode\\
 \hline
 $\alpha_{ep}$ & A scalar coefficient weighting the importance of the epistemic signal (set to $1$ in this work)\\
 \hline
 $\alpha_{in}$ & A scalar coefficient weighting the importance of the instrumental signal (set to $1$ in this work)\\
 \hline 
 $\gamma$ & The discount factor\\
 \hline 
 $A$ & The number of different continous motor actions to take at time $t$ (size of action space) \\
 \hline 
 $D$ & Dimensionality of input observation $\mathbf{o}_j$ \\
 \hline 
 $\Theta_a$ & The parameters/synapses that make up ActPC's motor-action circuit \\
 \hline 
 $\Theta_c$ & The parameters/synapses that make up ActPC's policy circuit \\
 \hline 
 $\Theta_g$ & The parameters/synapses that make up ActPC's generator/world model circuit \\
 \hline 
 $\Theta_p$ & The parameters/synapses that make up ActPC's prior preference circuit \\
 \hline 
 $\mathcal{M}$ & The uniform memory replay buffer (transition memories sampled uniformly) \\
 \hline 
 $\mathcal{M}^{actor}$ & The actor-specific memory replay buffer (transition memories sampled uniformly), also $\mathcal{M}^{a}$ \\
 \hline
\end{tabular}
\end{center}
\vspace{-0.5cm}
\end{table}

\noindent
\textbf{Derivation of State \& Weight Updates:} As mentioned in the main text, an NGC circuit, when it is presented with input stimuli (pair) $(\mathbf{x}_i, \mathbf{y}_i))$, minimizes an objective function known as total discrepancy (ToD). The ToD is formally:
\begin{align}
    \mathcal{L}(\Theta) = \sum^{L-1}_{\ell=0} \frac{1}{2} (|| \mathbf{z}^\ell - \mathbf{z}^\ell_{\mu} ||_2)^2
    = \sum^{L-1}_{\ell=0} \frac{1}{2} \sum_j \big( \mathbf{z}^\ell[j] - \mathbf{z}^\ell_{\mu}[j] \big)^2
    \label{eqn:tod}
\end{align}
where $\mathbf{z}^\ell[j]$ means that we extract the $j$-th element of vector $\mathbf{z}^\ell$. Since all of the latent states of the generative circuit are continuous, the updates will follow the form of the exact gradient, i.e., differentiation (which would permit the use of gradient descent), to optimize the latent variables and the synaptic parameters. Given this, the partial derivative of Equation \ref{eqn:tod} with respect to any layer of neural (state) activities $\mathbf{z}^\ell$ would be:
\begin{align}
     \frac{\partial \mathcal{L}(\Theta)}{\partial \mathbf{z}^\ell} &=  
                    \left( \frac{\partial \mathbf{z}^{\ell-1}_\mu}
                    {\partial \mathbf{z}^\ell} \cdot \Big( 
                    (\mathbf{z}^{\ell-1}-\mathbf{z}^{\ell-1}_\mu) \Big) \right)-
                    (\mathbf{z}^{\ell}-\mathbf{z}^{\ell}_\mu) \label{eqn:latent_update1}\\
    &= \Big[ (\mathbf{W}^\ell)^T \cdot  
                    (\mathbf{z}^{\ell-1}-\mathbf{z}^{\ell-1}_\mu) \Big] \odot \frac{\partial \phi^\ell(\mathbf{z}^\ell)}{\partial \mathbf{z}^\ell}  - (\mathbf{z}^{\ell}-\mathbf{z}^{\ell}_\mu) \label{eqn:latent_update2} \\
    &= (\mathbf{W}^\ell)^T \cdot (\mathbf{e}^{\ell-1}) \odot \frac{\partial \phi^\ell(\mathbf{z}^\ell)}{\partial \mathbf{z}^\ell}  - \mathbf{e}^\ell \label{eqn:latent_update3}
\end{align}
where we notice that the error neurons are derived directly from the ToD objective as well, i.e., $\mathbf{e}^\ell = \frac{\partial \partial \mathcal{L}(\Theta)}{\partial \mathbf{z}^\ell_\mu} = \mathbf{z}^\ell - \mathbf{z}^\ell_\mu$ (allowing us to write Equation \ref{eqn:latent_update2} in terms of error neurons as in Equation \ref{eqn:latent_update3}). 
Alternatively, by replacing the term $\frac{\partial \mathbf{z}^{\ell-1}}{\partial \mathbf{z}^\ell}$ with a learnable error matrix $\mathbf{E}^\ell$ instead, Equation \ref{eqn:latent_update2} can be simplified to the following:
\begin{align}
    \frac{\partial \mathcal{L}(\Theta)}{\partial \mathbf{z}^\ell} \approx \mathbf{d}^\ell =  \mathbf{E}^\ell \cdot \mathbf{e}^{\ell-1} - \mathbf{e}^\ell \label{eqn:z_l_delta}
\end{align}
which is a stable derivative-free perturbation $\mathbf{d}^\ell$ (so long as the activation function $\phi^\ell()$ is monotonically increasing)  to the latent neural activities (in \cite{ororbia2022neural} it was noted that the dampening effect of the activation derivative $\frac{\partial \phi^\ell(\mathbf{z}^\ell)}{\partial \mathbf{z}^\ell}$ can be approximated with a biologically-plausible dampening function if needed). The final update to the latent neural activities can then be performed using a gradient-ascent like operation, i.e.,  $\mathbf{z}^\ell \leftarrow \mathbf{z}^\ell + \beta \mathbf{d}^\ell$, which is the basic form of what was presented in the main paper -- a decay term $-\gamma \mathbf{z}^\ell$ (the ``leak'') is added to this update to smoothen out the update (i.e., a Gaussian prior placed over the latent state).

Deriving the updates to the synaptic generative parameters is also done in a similar fashion as above, i.e., by taking the gradient of ToD with respect to $\mathbf{W}^\ell$.
\begin{align}
    \frac{\partial \mathcal{L}(\Theta)}{\partial \mathbf{W}^\ell} \propto \Delta \mathbf{W}^\ell &=\frac{\partial \mathcal{L}(\Theta)}{\partial \mathbf{z}^\ell_\mu} \cdot  \left(\phi^{\ell+1}(\mathbf{z}^{\ell+1})\right)^T, \\
    &\mbox{where, } \mathbf{z}^\ell_\mu = \mathbf{W}^\ell \cdot \phi^{\ell+1}(\mathbf{z}^{\ell+1}) \nonumber \\ 
    &= (\mathbf{z}^\ell - \mathbf{z}^\ell_\mu) \cdot (\phi^{\ell + 1}( \mathbf{z}^{\ell+1}) )^T \\
    &= \mathbf{e}^\ell \cdot (\phi^{\ell + 1}( \mathbf{z}^{\ell+1}) )^T \mbox{.} \label{eqn:W_l_update}
\end{align}
If we are using $\mathbf{E}^\ell$ feedback/error matrices (as we do in this paper), we can leverage a simple Hebbian update $\Delta \mathbf{E}^\ell = \alpha \Big( \phi^{\ell+1}(\mathbf{z}^{\ell+1}) \cdot  (\mathbf{e}^\ell)^T \Big)$ \cite{ororbia2022neural} (which, if applied to $\mathbf{E}^\ell$ every time that Equation \ref{eqn:W_l_update} is applied to $\mathbf{W}^\ell$, allows $\mathbf{E}^\ell$ to come closer to the transpose of $\mathbf{W}^\ell$), such that solutions produced by both continuous function are within the fixed point, such that both solutions are stable. Much as was done for the states, synaptic weight matrices are updated via gradient ascent:  $\mathbf{W}^\ell = \mathbf{W}^\ell + \lambda \Delta \mathbf{W}^\ell$ and $\mathbf{E}^\ell = \mathbf{E}^\ell + \lambda \Delta \mathbf{E}^\ell$ ($\lambda$ is the learning rate/step size).

\noindent
\textbf{Relationship to Free Energy Minimization:} 
In ActPC, it is important to note that although the system's policy circuit is maximizing long-term discounted reward signals (produced jointly by both the prior and generator circuits), free energy \cite{friston2010free} is still being minimized as a result of the generator reducing its total discrepancy (when predicting future observations). 
Specifically, if one labels the ActPC's generative circuit as the ``transition model'', crucially noting that we set the encoder/decoder modules (needed to create a POMDP variant of our system) as identity matrices, 
and the policy circuit is the ``expected free energy estimation model'', which aligns ActPC with terminology used in work on deep active inference \cite{millidge2020deep,ueltzhoffer2018deep}), then an ActPC agent, much in the spirit of active inference, can be understood as seeking out observation/states that provide the greatest reduction in uncertainty given that our agent is most surprised when the epistemic value is highest.
As a result, (approximate) free energy will still be minimized because, upon choosing an action and transitioning to the next state/observation, ActPC's transition model will update its synapses to ensure that it is able to better predict the observation in the future. As the agent interacts with the environment multiple times, the generator will make fewer and fewer prediction errors, reducing its internal discrepancy and thus free energy, as it builds a more comprehensive picture of its world (assuming no sudden global changes or distribution shift).

\section*{ActPC: Discussion and Connections}





\begin{algorithm*}[!t]
\caption{The ActPC process under an environment $E$ for $N_e$ episodes (maximum episodic length is $T$). Note that we have color-coded portions of the algorithm as follows: \textcolor{blue}{blue} indicates signal calculation, \textcolor{orange}{orange} indicates memory storage operations, and \textcolor{red}{red} indicates synaptic updates (and the corresponding inference). See subsection (On ActPC Mechanics) for explanation of sub-routines $f_{proj}(.)$,  $f_{inf}(.)$, and $\text{AdjustSynapses}(.)$. We also provide detail about each symbols in our notation table for reader's convention.
}
\label{algo:actpc_process}
\begin{algorithmic}[1]
   \State {\bfseries Input:} environment $E$, actor $\Theta_a$, policy $\Theta_c$, generator $\Theta_g$, prior $\Theta_p$, replay memory $\mathcal{M}$, actor memory $\mathcal{M}^{actor}$
   \State {\bfseries Constants:} $N_e$, $T$, $\alpha_{ep}$, $\alpha_{in}$, $\gamma$, $A$
   \Function{Simulate}{$E, N_e, T, \Theta_a, \Theta_c, \Theta_g, \Theta_p, \mathcal{M}, \mathcal{M}^{actor}, \alpha_{ep}, \alpha_{in}, \gamma$}
        \State $r^{ep}_{max} = 1$, $r^{in}_{max} = 1$
        \For{$n = 1$ to $N_e$}
            \State $\mathbf{o}_t \leftarrow \mathbf{o}_0$ from $E$, \; $r_E = 0$ \Comment Sample initial observation from environment
            \For{$t = 1$ to $T$}
              \State $\mathbf{a}_t = f_{proj}(\mathbf{o}_t; \Theta_a)$ \Comment Get action $\mathbf{a}_t$ from actor circuit
              \LineComment Get next state/observation from environment \& compute component reward signals
              \State $( r_t, \mathbf{o}_{t+1} ) \leftarrow E(\mathbf{a}_t)$, \; $(\Lambda_g, \mathcal{E}_g) = f_{inf}([\mathbf{a}_t, \mathbf{o}_t], \mathbf{o}_{t+1}; \Theta_g)$, \; $(\Lambda_p, \mathcal{E}_p) = f_{inf}(\mathbf{o}_t, \mathbf{o}_{t+1}; \Theta_p)$
              \State $r_E \leftarrow r_E + r_t$ \Comment Track cumulative episodic sparse reward
              \color{blue}
              \State $\hat{r}^{ep}_t = \sum_\ell ||\mathcal{E}_g[\ell]||^2_2$, \; $r^{ep}_{max} = \max(\hat{r}^{ep}_t, r^{ep}_{max})$, \; $r^{ep}_t \leftarrow  \frac{\hat{r}^{ep}_t}{ r^{ep}_{max} }$ \Comment Calculate epistemic signal
              \State $\hat{r}^{in}_t = \sum_\ell ||\mathcal{E}_p[\ell]||^2_2$, \; $r^{in}_{max} = \max(\hat{r}^{in}_t, r^{in}_{max})$, \; $r^{in}_t \leftarrow  -\frac{\hat{r}^{in}_t}{ r^{in}_{max} }$ \Comment Calculate instrumental signal
              \State $r_t = \alpha_{ep} r^{ep}_t + \alpha_{in} r^{in}_t$ \Comment Calculate internal reward signal
              \color{orange}
              \LineComment Store current transition and sample from memory (via combined experience replay \cite{zhang2017deeper})
              \State Store $(\mathbf{o}_t, \mathbf{a}_t, r_t, \mathbf{o}_{t+1})$ in $\mathcal{M}$
              \State $(\mathbf{o}_t, \mathbf{a}_t, r_t, \mathbf{o}_{t+1}) \cup (\mathbf{o}_j, \mathbf{a}_j, r_j, \mathbf{o}_{j+1}) \sim \mathcal{M}$ \Comment{Sample mini-batch \& combine w/ current transition}
              \color{red}
              \LineComment Update motor-action circuit $\Theta_a$
              \State $(\Lambda_a, \mathcal{E}_a, \Lambda_c, \mathcal{E}_c) = f_{inf}([\mathbf{a}_t, \mathbf{o}_t], \emptyset; \Theta_c \cup \Theta_a), \mbox{ with } \mathbf{e}_0 = -\mathbf{1}_a/A$, \; $\Theta_a \leftarrow \text{AdjustSynapses}{\Lambda_a, \mathcal{E}_a, \Theta_a}$ 
              \LineComment Update policy circuit $\Theta_a$
              \State $ \mathbf{c}_j = f_{proj}([\mathbf{a}_j, \mathbf{o}_{j+1}]; \Theta_c)$, \; 
              $\Big( \mathbf{o}_t \mbox{ is terminal} \rightarrow \mathbf{t}_t = r_j \mathbf{1}_a \Big) \land \Big( \mathbf{o}_t \mbox{ is not terminal} \rightarrow \mathbf{t}_t = r_j + \gamma \mathbf{c}_j \Big)$
              \State $(\Lambda_c, \mathcal{E}_c) = f_{inf}([\mathbf{a}_j, \mathbf{o}_j], \mathbf{t}_j; \Theta_c)$, \; $\Theta_c \leftarrow \text{AdjustSynapses}{\Lambda_c, \mathcal{E}_c, \Theta_c}$ 
              \LineComment Update generator circuit $\Theta_a$
              \State $(\Lambda_g, \mathcal{E}_g) = f_{inf}([\mathbf{a}_t, \mathbf{o}_t], \mathbf{o}_{t+1}; \Theta_g)$, \; $\Theta_g \leftarrow \text{AdjustSynapses}{\Lambda_g, \mathcal{E}_g, \Theta_g}$
              \LineComment Perform a step of motor-action refresh
              \State $(\mathbf{o}_j, \mathbf{a}_j) \sim \mathcal{M}^{actor}$, \; $(\Lambda_a, \mathcal{E}_a) = f_{inf}(\mathbf{o}_j, \mathbf{a}_j; \Theta_g)$, \; $\Theta_g \leftarrow \text{AdjustSynapses}{\Lambda_a, \mathcal{E}_a, \Theta_a}$
              \color{black}
              \State $\mathbf{o}_t \leftarrow \mathbf{o}_{t+1}$
            \EndFor
            \color{orange}
            \LineComment Store episode $n$'s transitions into motor-action replay buffer $\mathcal{M}^{actor}$
            \State $ r^{max}_E \leftarrow \mathcal{M}^{actor}$, \; $\mbox{if } r_E > 0 \mbox{ and } r_E \geq r^{max}_E \mbox{ then store } \{(\mathbf{o}_i, \mathbf{a}_i, r_i, \mathbf{o}_{i+1})\}^{i=T}_{i=0} \mbox{ in } \mathcal{M}^{actor}$
        \EndFor
   \EndFunction
\end{algorithmic}
\end{algorithm*}

\subsection*{On ActPC Mechanics}
\label{ngc_mechanics}



\subsubsection*{Algorithmic Depiction}

In Algorithm \ref{algo:actpc_process}, we depict the full ActPC inference and learning process across $N_e$ episodes of maximum length $T$ (note that $T$ could be variable).\footnote{Please refer to the main paper for the neuronal-level equations that back the ancestral projection, iterative inference, and synaptic update routines that we represent here with short-hand in the pseudocode in this appendix.}
\textbf{Projection:} We represent, as stated in the main paper, ancestral projection through any given NGC circuit as $f_{proj}(\mathbf{x}; \Theta)$, where $\mathbf{x}$ is the input to the circuit (e.g., $\mathbf{o}_t$, or concatenated inputs such as $[\mathbf{a}_t, \mathbf{o}_t]$), $\Theta$ is the circuit synaptic parameters ($\Theta_a$ refers to the motor-action synapses, $\Theta_c$ refers to the policy synapses, $\Theta_g$ refers to the generator synapses, and $\Theta_p$ refers to the prior synapses).
\textbf{Iterative Inference:} We represent a circuit's iterative inference/state-update process as $f_{inf}(\mathbf{x}, \mathbf{y}; \Theta)$ where $\mathbf{x}$ is the input to the circuit and $\mathbf{y}$ is the intended target for the output/bottom-most layer $\mathbf{z}^0$, e.g., $f_{inf}([\mathbf{a}_t, \mathbf{o}_t], \mathbf{t}_t; \Theta_c)$ means that we provide the inputs $[\mathbf{a}_t, \mathbf{o}_t]$ and the target $\mathbf{t}_t$ to the policy circuit $\Theta_c$. Critically note that this inference procedure returns two sets of statistics, the inferred values of the latent neural states stored in $\Lambda$ and the current error neuron values stored in $\mathcal{E},$ -- for example, $(\Lambda_c, \mathcal{E}_c) = f_{inf}([\mathbf{a}_t, \mathbf{o}_t], \mathbf{t}_t; \Theta_c)$ return statistics the latent state and error neuron values for the policy circuit $\Theta_c$.
\textbf{Synaptic Updating:} Finally, we represent the local Hebbian adjustment to a particular circuit's synapses as $\Call{AdjustSynapses}{\Lambda, \mathcal{E}, \Theta}$, which returns a new set of parameters $\Theta'$ -- for example, $\Call{AdjustSynapses}{\Lambda_a, \mathcal{E}_a, \Theta_a}$ means that we would use the statistics $(\Lambda_a, \mathcal{E}_a)$ to physically adjust the synaptic parameters of the motor-action circuit $\Theta_a$. 
Note that, in Algorithm \ref{algo:actpc_process}, $\leftarrow$ indicates variable overriding, e.g., $\Theta_g \leftarrow \Theta'$ means we override the original generator synaptic parameters $\Theta_g$ with $\Theta'_g$.

The final ingredient needed for understanding Algorithm \ref{algo:actpc_process} is the meaning of the first portion of Line 19 (which was also described in the main paper). Specifically, this portion of pseudocode is: ``$(\Lambda_a, \mathcal{E}_a, \Lambda_c, \mathcal{E}_c) = f_{inf}([\mathbf{a}_t, \mathbf{o}_t], \emptyset; \Theta_c \cup \Theta_a), \mbox{ with } \mathbf{e}_0 = -\mathbf{1}_a/A$''. Specifically, this line refers to the coupling between the motor-action circuit and the policy circuit that is needed in order to compute the synaptic updates for the motor-action parameters $\Theta_a$. No target vector is provided to this routine, hence the empty set $\emptyset$, and the error neurons of the policy circuit are switched to the form $\mathbf{e}_0 = -\mathbf{1}_a/A$ (by setting $u_a = 1$) -- observe that we use the notation $\Theta_c \cup \Theta_a$ to indicate that we are treating the motor-action and policy circuits as one single module/larger circuit and running iterative inference over their union. 
Note that this (coupled-circuit) process returns four sets of statistics, i.e., latent state and error neuron values $(\Lambda_a, \mathcal{E}_a)$ for the motor-action circuit and latent state and error neuron values $(\Lambda_c, \mathcal{E}_c)$ for the policy circuit (the latter of which we never use and simply discard).

Finally, we represent the individual memory replay buffers used to train the ActPC system as $\mathcal{M}$ for the uniform replay buffer and $\mathcal{M}^{actor}$ for the motor-action specific buffer. In Line 26 of Algorithm \ref{algo:actpc_process}, we highlight that we store the maximum best cumulative sparse reward seen so far for any given episode and use this as a filter/condition for storing successful episodes in the motor-action module's buffer.

\subsection*{ActPC Joint Objective}

In effect, the process shown in Algorithm \ref{algo:actpc_process} depicts how the ActPC adapts itself dynamically to transitions sampled from the environment. If we represent the policy circuit as $Q(\mathbf{o}_t,\mathbf{a}_t|\Theta_c)$ and the actor circuit as $\mu(\mathbf{s}_t|\Theta_a)$, then the objective that the ActPC agent $\mathcal{A}$ optimizes can be decomposed into the following sub-objectives: 
\begin{enumerate}[noitemsep,nolistsep]
    \item the generator circuit optimizes its own ToD (attempting to learn an internal world model of $E$ -- Equation \ref{eqn:tod}),
    \item the policy circuit optimizes its ToD which also optimizes (if we focus on the output layer neurons $\mathbf{e}^0$):
\begin{align}
    \mathcal{L}(\Theta_c) &= \mathbb{E}_{\mathbf{o}_t \sim E, \mathbf{a}_t \sim \pi, r_t \sim \mathcal{A}}\Big[ \big( Q(\mathbf{o}_t,\mathbf{a}_t|\Theta_c) - \mathbf{t}_t \big)^2 \Big] \\
    &\mbox{where } \mathbf{t}_t = r_t + \gamma Q\big(\mathbf{o}_{t+1},\mu(\mathbf{a}_{t+1})|\Theta_c)\big) \nonumber 
\end{align}, 
    \item the motor-action circuit optimizes its own ToD that also optimizes (again, if we focus on the output layer's neurons $\mathbf{e}^0$, or the error units associated with the output units):
    \begin{align}
        \nabla_{\Theta_a} J = \mathbb{E}_{\mathbf{o}_t \sim E}&\Big[\nabla Q(\mathbf{o},\mathbf{a}|\Theta_c)|_{\mathbf{o}=\mathbf{o}_t,\mathbf{a}=\mu(\mathbf{o}_t)} * \\
        &\nabla_{\Theta_a} \mu(\mathbf{o}|\Theta_a)|_{\mathbf{o}=\mathbf{o}_t} \Big] \nonumber \\
        &+ \lambda \mathbb{E}_{\mathbf{o}_t \sim E, \mathbf{a}_t \sim \mathcal{M}^{a}}\Big[ \big( \mu(\mathbf{o}_t|\Theta_a) - \mathbf{a}_t \big)^2 \Big] \nonumber 
    \end{align}
    where $\mathcal{M}^{a}$ is the actor refresh memory and the term $J = \mathbb{E}_{r_i,\mathbf{o}_i \sim E, \mathbf{a}_i \sim \pi}\big[ r_1\big]$ is the start distribution. 
    The above equation (if $\lambda = 0$) represents the gradient of the policy's performance, or the policy gradient \cite{silver2014deterministic}. Finally, notice the second term in the above equation, highlighting that the actor refresh step in Algorithm \ref{algo:actpc_process} introduces a second term that focuses on optimizing the ToD for the actor circuit alone.
\end{enumerate}
As a result, one can in effect view the ActPC as optimizing a set of ToD (or approximate free energy) objectives that approximately follow the policy gradient, highlighting that our motor-action and policy circuits jointly optimize the actor-critic objective based on the deterministic policy gradient (DPG) approach/formulation \cite{silver2014deterministic}.

\subsection*{Motivations}
Given that ActPC is motivated by active inference \cite{friston2011action}, the intuition behind our approach is that an agent reduces the divergence between its model of the world and the actual world by either adapting the internal model so that it is more predictive of sampled observations (motivating the agent to seek highly surprising states) or changing its observations such that they better align with the internal world model (through choosing actions/interacting with the environment). This process is also influenced by the prior circuit, acting as the agent's prior preference \cite{friston2009reinforcement,tschantz2020learning}, placing a pressure on the agent to seek goal states. 
The epistemic signal connects ActPC to intrinsic curiosity-based methodology, e.g., exploration bonuses \cite{wu2016training}, while the generator connects our work to the growing interest in model-based RL that integrates learned world models, e.g., dreamer \cite{hafner2019dream}, plan2explore \cite{sekar2020planning}. In essence, one could view ActPC as a neurobiologically-plausible instantiation of model-based RL. 
Many other bio-inspired algorithms, such as equilibrium propagation \cite{scellier2017equilibrium} or contrastive Hebbian learning \cite{movellan1991contrastive}, do not scale well to RL problems due to expensive inference. 

Crucially, the ActPC framework prescribes a coordinated interaction of its policy, actor, prior, and generator modules. The motor-action circuit chooses which action to take next, the policy estimates the future discounted reward-like signals associated with an action, and the prior and generator circuits attempt to guess the (next) state of the environment to produce the agent's internal reward signal. Upon taking action $\mathbf{a}_t$, the generator's prediction is corrected using sensory samples drawn from environment $E$, progressively improving the agent's internal impression of $E$. The inability of the generator to accurately predict the incoming $\mathbf{o}_{t+1}$ creates a strong guide for the agent to explore its world:
\begin{align}
    \widehat{r}^{ep}_t &= \sum_j (\mathbf{e}^0 )^2_{j,1} + \sum_j (\mathbf{e}^1 )^2_{j,1} + \sum_j (\mathbf{e}^2 )^2_{j,1} \\
    r^{ep}_t &= \widehat{r}^{ep}_t / (r^{ep}_{max}) \quad \mbox{where } r^{ep}_{max} = \max(\widehat{r}^{ep}_1, \widehat{r}^{ep}_2, ..., \widehat{r}^{ep}_t) \mbox{.} \label{eqn:epistemic_dup}
\end{align}
Ultimately, ActPC aims to reduce the long-term surprisal experienced by its generator circuit which will require the motor-action and policy circuits to extract effective plans to reach goal states. 

\subsection*{On the Prior Preference Circuit}

During our experimental simulations, we first train the prior preference circuit (again, running the setting process in Equation \ref{eqn:state_update} and adjusting synapses with the local rules in Equations \ref{eqn:predictor_update}-\ref{eqn:error_update}) on a small set of demonstration samples. Then, during simulation, we fix its synaptic updates (setting its learning rate to zero) for the rest of the ActPC agent's simulation, which means that, at every time step, the prior aims to predict the next observation given its knowledge of preferred action trajectories. In other words, the prior circuit tells the motor-action and policy circuits what observation/state it would have preferred/desired to see if the aim is to reach a task/problem goal state.

The prior preference circuit produces the instrumental signal needed to drive the agent towards goals as follow:
\begin{align}
    \widehat{r}^{in}_t &= \sum_j (\mathbf{e}^0 )^2_{j,1} + \sum_j (\mathbf{e}^1 )^2_{j,1} + \sum_j (\mathbf{e}^2 )^2_{j,1} \\
    r^{in}_t &= -\widehat{r}^{in}_t / (r^{in}_{max}) \quad \mbox{where } r^{in}_{max} = \max(\widehat{r}^{in}_1, \widehat{r}^{in}_2, ...,\widehat{r}^{in}_t) \label{eqn:instrumental_dup}
\end{align}
where we observe that we dynamically track the maximum value in order to normalize the instrumental signal to the range of $[0,1]$.

Note that for all Mujoco agent models, the prior circuit was set to contain two hidden layers configured as $[512,256]$ employing linear rectifier (relu) activation functions. The robosuite agent models, on the other hand, had prior circuits with hidden layers configured as $[512,512]$ with relu6 activation functions (relu but with an upper bound of $6$ imposed each neuron's value).

\subsection*{On Biological Connections}

The fundamental NGC circuits that compose our ActPC agent align with the various predictive processing computational models that have been proposed to explain brain function \cite{rao1999predictive,friston2005theory,bastos2012canonical}. Desirably, this provides some grounding of our model in computational cognitive neuroscience by connecting it to a prominent Bayesian brain theory as well as with established general principles of neurophysiology, e.g., NGC combines diffuse inhibitory feedback connections with driving feedforward excitatory connections. In addition, the synaptic adjustments computed by any NGC circuit are local, corresponding to, if we include the factor $1/2\beta_e$ (to replace the normally learnable precision matrix of \cite{ororbia2020neural}), a three-factor (error) Hebbian update rule. Nevertheless, there are many elements of the NGC component circuit that preclude it from serving as a complete and proper computational model of biological neural circuitry, e.g., neurons are communicating with real-values instead of spikes \cite{wacongne2012neuronal}, synapses are currently allowed to be negative and positive, etc.. However, the backprop-free ActPC system represents a step forward towards agent designs that facilitate scalable simulations of neuro-biologically plausible computation that also generalizes well on complicated problems such as those found in robotics. 
Practically, one could take advantage of the parallelism afforded by high performance computing to simulate very large, non-differentiable neural circuits given that our NGC circuits are naturally layer-wise parallel and do not suffer from the forward and backward-locking problems that plague backprop \cite{jaderberg2017decoupled}.
Finally, the ActPC agent framewrok contributes to the effort to create more biologically-motivated models and update rules either based on or for reinforcement learning \cite{mazzoni1991more,alexander2018frontal,yamakawa2020attentional} and motor control.

\section*{Baselines and Discussion} 

In this section, we describe, in further detail, the baseline RL methods that we compared against for both the Mujoco and robosuite robotic control tasks. 
Please see Tables \ref{table:configs_actpc}-\ref{table:configs_ddpg_demo} for hyper-parameter configurations/settings for all baselines and ActPC for all benchmark problems. Further note that all the baselines, ranging from DDPG to LOGO had access to and exploited internally demonstration data as this was found to be necessary to obtain meaningful performance.

\textbf{DDPG and DDPG-Demo:}: Deep deterministic policy gradients (DDPG) \cite{lillicrap2015continuous}, ) is a model-free RL algorithm for continuous action spaces. In DDPG there are two networks, an actor and a critic, and the job of the critic is to approximate actor’s action-value function. DDPG-Demo, the model we compare to in the robosuite experiments, is a powerful extension to the DDPG algorithm that leverages additional demonstration data (note that without it, the DDPG model failed to perform in any meaningful manner on all the robotic control tasks).

\noindent
\textbf{TD3:} Twin-delayed DDPG (TD3) \cite{td3_18} is an approach in which the actor-critic component of the agent considers the interplay between function approximation error in
both the policy and value updates. We use three fully connected layers with up to $512$ units for the hidden layers for both the actor and critic. We use the linear rectifier (relu) as the activation function in all portions of the model, except the final actor layer which uses the hyperbolic tangent as the activation (and the identity for the output layer of the critic), a batch size of $1024$, a discount factor $\gamma = 0.98$, a buffer size $1e6$, and we use Adam with a learning rate of $2e-5$ to update the model parameters \cite{kingma2014adam}, 

\noindent
\textbf{HER:} In hindsight experience replay (HER) \cite{andrychowicz2017hindsight}, the critical insight is that, even in failed rollouts with no reward from RL environment, the agent can transform the resultant trajectories into successful ones by assuming that a state that it observed in the given rollout was the actual goal (state). 
For every episode that the RL agent experiences, we store these statistics into the replay buffer twice: once for the original goal pursued and once the other for the final state achieved in the episode, assuming the agent intended to reach this last state from the very beginning. We use three fully-connected layers each with $[256-512]$ hidden units for both actor and critic with the relu activation function (except for the output layers, as in TD3), a batch size of $2048$, a discount factor $\gamma = 0.98$, a buffer size of $1e6$, and optimize model parameters using Adam with a learning rate of $3e-5$. 

\noindent
\textbf{LOGO:} In learning online with guidance offline (LOGO) \cite{logo21}, at each iteration of processing, the algorithm consists of two steps, namely a policy improvement step and a policy guidance step. 
\textbf{Step 1:} Policy Improvement: In this step, the LOGO algorithm performs a one-step policy improvement/adjustment using the Trust Region Policy Optimization (TRPO) approach \cite{schulman2015trust}, and 
\textbf{Step 2:} Policy Guidance: the TRPO approach used in Step 1 works stably in dense reward settings but fails when applied on sparse rewards. Hence, to overcome this problem, LOGO leverages offline demonstration data to provide enhanced policy guidance such that the model moves closer to a good behavior policy.
In our implementation, we use two fully connected layers each with up to $512$ hidden units with tanh activation to parameterize the policy and values. We use Adam with a learning rate of $2e-5$ to updates model weights with a batch size of $2048$. We keep LOGO specific hyper-parameters consistent with the ones reported in the original source paper (as we found in preliminary experimentation that these worked best for the control problems studied). For offline demonstration data, we trained dense reward models for only $2000$ episodes and collected the statistics to create demonstration buffer/pool.

\section*{Related Work} 
\label{sec:lit_review}

\noindent
\textbf{Learning from Demonstration}: The key idea behind Learning from Demonstration (LfD) is to use demonstration (demo) data to provide a better prior to aid RL-based approaches. One widely used method to accelerate learning is to add exploration data into the replay buffer which is then tuned using a priority replay scheme \cite{nair2018overcoming, hester2018deep, vecerik2017leveraging}. Another approach that improves priority replay is combining policy gradients with demo data by combining behavior cloning with online RL learning \cite{rajeswaran2017learning}. Another similar method is leveraging a large amount of offline data with associated rewards to facilitate online RL better using AWAC \cite{nair2020accelerating}. Agents that are closer to our work that work well with sparse rewards include LOGO \cite{logo21} and PofD \cite{kang2018policy}; they require demo data to be extracted from a dense reward model. However, PofD modifies the reward by considering the weighted average of the sparse reward from the online agent and the implicit reward received from dense offline model. Similar to LOGO, ActPC do not modify/depend on any dense/shaped reward functions and purely works with sparse reward signals.
 
\noindent
\textbf{Offline Learning}: Recently there have been a few exciting works \cite{wu2019behavior, siegel2020, kumar2019, fujimoto2019} on learning policy using offline data as opposed to online learning. In contrast, ActPC is a purely online RL system.

\noindent
\textbf{Bio-inspired RL}: 
Hebbian RL \cite{najarro2020meta} is a meta-learning RL scheme that relies on an evolutionary strategy to adapt its core model parameters (e.g., the Hebbian coefficients). In contrast, the ActPC framework focuses on crafting an online, local scheme that does not depend on evolution.
Furthermore, \cite{najarro2020meta} focused on the challenge of continual RL, whereas this study was focused on demonstrating the viability of a backprop-free framework for handling RL robotic control problems. Nevertheless, it would be useful to investigate how ActPC would operate in a continual RL setting. 
In other bio-inspired approaches, the learning process is decomposed based on the source of reward, which serves as a goal for any given subtask \cite{zhou2004biologically}. In addition, an artificial emotion indication (AEI) is assigned for each subtask, where the AEI is responsible for predicting the reward component associated with any particular given subtask.
Another approach \cite{weibio-inspired} demonstrated that the backprop algorithm created issues for the learning process in modern-day neural-based RL models, particularly for those RL models meant to obtain desired solutions in a minimum amount of time and crafted a bio-inspired alternative to tackle this issue. Specifically in this work, the network weights undergo spontaneous fluctuations and a reward signal is responsible for modulating the center and amplitude of these fluctuations. This ensures faster and enhanced convergence such that the RL network model acquires a desired behavior. (We note that such an approach could be complementary to the ActPC formulation.)

Although a myriad of biologically-inspired algorithms have been proposed to replace backprop \cite{movellan1991contrastive,o1996biologically,lee2015difference,lillicrap2016random,scellier2017equilibrium,guerguiev2017towards,whittington2017approximation,ororbia2018biologically,ororbia2020large} over the past several years, research has largely focused on developing methods in the context of classification with some exceptions, e.g., generative modeling and temporal data \cite{wiseman2017training,ororbia2020continual,manchev2020target,ororbia2020neural}. Although any backprop-alternative could be, in principle, reformulated for RL, many such algorithms either require very long chains of computation to simulate their inference, such as contrastive Hebbian learning \cite{movellan1991contrastive} and equilibrium propagation \cite{scellier2017equilibrium}, or do not offer complete approaches to resolving all biological criticisms of backprop, for example, representation alignment \cite{ororbia2018biologically}, feedback alignment \cite{lillicrap2016random}, and target propagation \cite{lee2015difference} algorithms do not resolve the update and backwards locking problems \cite{jaderberg2017decoupled} while approaches such as \cite{whittington2017approximation} do not resolve the weight transport problem and require activation function derivatives. In contrast, NGC has been shown to offer one complete approach to resolving most of backprop's key criticisms and does not require multiple expensive inference phases \cite{ororbia2020neural}. Furthermore, NGC has already been empirically shown to work well in developing temporal generative models without unfolding the computation graph through time (as is critical in backprop through time) \cite{ororbia2020continual,ororbia2020neural}. Since the motivation behind this work is to develop a flexible agent that embodies active inference yet conducts inference and synaptic adjustment in a neurocognitively-plausible fashion, NGC serves as an important foundation to craft the fundamental agent circuits from.

\noindent
\textbf{Imitation Learning:} In imitation learning (IL), the agent's goal is to imitate an (expert) policy using the demonstration data generated by that policy. One of the widely-used and practical approaches is based on the simple concept of behavior cloning (BC), in which the expert policy is estimated using traditional supervised learning leveraging samples of demonstration data. However, BC-based approaches struggle whenever there is a distribution shift \cite{ross2011reduction}. Other approaches take the form of inverse RL (IRL), in which the agent tries to solve a forward RL problem using the reward function estimated from the data. 
There are a few approaches that avoid reward estimation. For instance, generative adversarial imitation learning (GAIL) \cite{ho2016generative} takes a distribution matching approach and provides an implicit reward for the agent using a discriminator. Most IRL algorithms do not use reward feedback from the environment and hence are restricted to the performance of the policy that is extracted/generated the demonstration data. Our approach differs from pure IRL; we leverage online RL with (sparse) reward feedback to drive efficient learning.

\noindent
\textbf{Model-Based RL} Since ActPC employs and adapts a dynamic generative model to produce the necessary signals to drive its curiosity/exploration, our agent also contributes to the work in improving model-based RL through the use of world models \cite{ha2018recurrent,nagabandi2018neural,chua2018deep,hafner2019learning}. However, world models in modern-day, model-based RL systems are learned with backprop whereas our ActPC agent uses the same parallel neural processing and gradient-free weight updating throughout (for all its individual circuits including the generative world model), further obviating the need for learning the entire system in pieces or using evolution to learn an action model \cite{ha2018recurrent}.

\noindent
\textbf{Intrinsic Curiosity} Key to our overall ActPC agent design is the notion of novelty or surprise \cite{barto2013novelty}, which is what provides our neural system (through the epistemic signal $r^{ep}_t$) with a means to explore an environment beyond a uniform random sampling scheme. This connects our ActPC framework to the family of RL models that have been designed to generate intrinsic reward signals \cite{schmidhuber1991possibility,storck1995reinforcement,singh2004intrinsically,oudeyer2009intrinsic}, which are inspired by the psychological concept of curiosity inherent to human agents \cite{ryan2000intrinsic,silvia2012curiosity}. Crucially, curiosity provides an agent with the means to acquire new skills that might prove useful for maximizing rewards in downstream tasks. 
We note that there are many other forms of intrinsic reward signals such as those based on policy-entropy \cite{rawlik2013probabilistic,haarnoja2018soft}, information gain \cite{houthooft2016vime,shyam2019model}, prediction error \cite{pathak2017curiosity,burda2018exploration},
state entropy \cite{lee2019efficient}, state uncertainty \cite{o2018uncertainty}, and empowerment \cite{leibfried2019unified,mohamed2015variational}. 

\noindent
\textbf{Active Inference} Finally, our ActPC agent framework offers a simple predictive processing interpretation of active inference and the more general theoretical framework known as planning-as-inference.
Planning-as-inference (PAI) \cite{botvinick2012planning} puts forth the view that a decision-making agent utilizes an internal cognitive model to represent its future as the joint probability distribution over actions, (outcome) states, and rewards. 
Active inference \cite{friston2017active_curosity}, offering one means of implementing PAI, suggests that agents choose actions by maximizing the evidence for an internal (generative) model that is further biased towards the agent’s preferences. While this work focused on crafting a (fixed) neural circuit that generates scalar signals to drive goal-orienting behavior, which would likely be a component of the error function that is embodied in the firing rates of dopamine neurons \cite{montague1994predictive,montague1996framework,schultz1997neural}, we remark that one could instead encode goal states or other task-orienting functions that could induce more complex behavior or facilitate longer-term planning in ActPC. 

\begin{table*}[!t]
\begin{center}
\begin{tabular}{l|lll|lll|lll}
 & \multicolumn{3}{c}{\textbf{Reacher}} & \multicolumn{3}{|c}{\textbf{Half-Cheetah}} & \multicolumn{3}{|c}{\textbf{Walker}}\\
 & \textbf{Actor} & \textbf{Policy} & \textbf{Generator} & \textbf{Actor} & \textbf{Policy} & \textbf{Generator} & \textbf{Actor} & \textbf{Policy} & \textbf{Generator}\\
 \hline 
 $\phi(\cdot)$ & ReLU6  &  ReLU6 & ReLU6  &  ReLU6 & ReLU6  &  ReLU6 &  ReLU6 & ReLU6  &  ReLU6 \\
 Dims & $[256, 256]$ & $[256, 256]$  & $[256,256]$ & $[256,256]$ & $[256,256]$ & $[256,256]$ & $[256,256]$ & $[256,256]$ & $[256,256]$ \\
 Rule & Adam & Adam & Adam & Adam & Adam & Adam & Adam & Adam & Adam   \\
 $\eta$ & $0.0003$ & $0.0003$ & $0.0003$ & $0.0004$ & $0.0004$ & $0.0004$ & $0.003$ & $0.0003$ & $0.0003$\\
 $\tau$ & $0.005$ & $0.005$ & --  & $0.05$ & $0.05$ & -- & $0.15$ & $0.15$ & -- \\
 $\gamma$ & -- & $0.99$ & -- &  -- & $0.99$ & -- &  -- & $0.99$ & -- \\
 $N_{mem}$ & $10^6$ & $10^6$ & $10^6$ & $10^6$ & $10^6$ & $10^6$ & $10^6$ & $10^6$ & $10^6$ \\
 $N^{demo}_{mem}$ & $10^5$ & $10^5$ & $10^5$ & $10^5$ & $10^5$ & $10^5$ & $10^5$ & $10^5$ & $10^5$ \\
 $N^{actor}_{mem}$ & $10^5$ & $10^5$ & $10^5$ & $10^5$ & $10^5$ & $10^5$ & $10^5$ & $10^5$ & $10^5$ \\
 $N_{batch}$ & $256$ & $256$ & $128$ & $128$ & $256$ & $256$ & $128$ & $256$ & $256$ \\
 \hline
\end{tabular}
\caption{The meta-parameter configurations used for ActPC.}
\label{table:configs_actpc}
\begin{tabular}{l|ll|ll|ll}
 & \multicolumn{2}{c}{\textbf{Reacher}} & \multicolumn{2}{|c}{\textbf{Half-Cheetah}} & \multicolumn{2}{|c}{\textbf{Walker}}\\
 & \textbf{Actor} & \textbf{Critic}  & \textbf{Actor} & \textbf{Critic}  & \textbf{Actor} & \textbf{Critic} \\
 \hline 
 $\phi(\cdot)$ & ReLU  &  ReLU &  ReLU&  ReLU&  ReLU & ReLU   \\
 Dims & $[512, 256]$ & $[512, 256]$ & $[512,256]$ & $[512,256]$ & $[512,256]$  & $[512,256]$  \\
 Rule & AdamW & Adam & Adam & Adam & Adam & Adam   \\
 $\eta$ & $0.0001$ & $0.0003$  & $0.0005$ & $0.0004$ & $0.0005$  & $0.0002$  \\
 $\tau$ & $0.005$ & $0.005$  & $0.005$ & $0.005$ & $0.005$  & $0.005$  \\
 $policy_{noise}$ & $--$ & $0.3$  & $--$ & $0.3$ & $--$  & $0.3$  \\
 $policy_{freq}$ & $--$ & $5$  & $--$ & $5$ & $--$  & $5$  \\
 $expl_{noise}$ & $0.1$ & $0.1$  & $0.1$ & $0.1$ & $0.1$  & $0.1$  \\
 $\gamma$ & $--$ & $0.991$  & $--$ & $0.992$  &  $--$  & $0.992$  \\
 $N_{mem}$ & $10^6$ & $10^6$  & $10^6$ & $10^6$ & $10^6$ & $10^6$  \\
 $N^{demo}_{mem}$ & $10^5$ & $10^5$ & $10^5$ & $10^5$ & $10^5$ & $10^5$ \\
 $N_{batch}$ & $1024$ & $1024$  & $1024$ & $1024$ & $1024$  & $1024$  \\
 \hline
\end{tabular}
\caption{The meta-parameter configurations used for TD3.}
\label{table:configs_td3}

\begin{tabular}{l|ll|ll|ll}
 & \multicolumn{2}{c}{\textbf{Reacher}} & \multicolumn{2}{|c}{\textbf{Half-Cheetah}} & \multicolumn{2}{|c}{\textbf{Walker}}\\
 & \textbf{Actor} & \textbf{Critic}  & \textbf{Actor} & \textbf{Critic}  & \textbf{Actor} & \textbf{Critic} \\
 \hline 
 $\phi(\cdot)$ & ReLU  &  ReLU &  ReLU&  ReLU&  ReLU & ReLU   \\
 Dims & $[256, 256]$ & $[256, 256]$ & $[256,256]$ & $[256,256]$ & $[256,256]$  & $[256,256]$  \\
 Rule & AdamW & Adam & Adam & Adam & Adam & Adam   \\
 $\eta$ & $0.0001$ & $0.00003$  & $0.0005$ & $0.00004$ & $0.0005$  & $0.00002$  \\
 $\epsilon$ & $1.0$ & $1.0$  & $1.0$ & $1.0$  &  $1.0$  & $1.0$  \\
 $\epsilon_{decay}$ & $0.9999$ & $0.9999$  & $0.9999$ & $0.9999$  &  $0.9999$  & $0.9999$  \\
 $\epsilon_{min}$ & $0.32$ & $0.32$  & $0.32$ & $0.32$  &  $0.32$  & $0.32$  \\
 $mem_{\alpha}$ & $0.6$ & $0.6$  & $0.6$ & $0.6$  &  $0.6$  & $0.6$  \\
 $mem_{\epsilon}$ & $0.01$ & $0.01$  & $0.01$ & $0.01$  &  $0.01$  & $0.01$  \\
 $mem_{\beta}$ & $0.4$ & $0.4$  & $0.4$ & $0.4$  &  $0.4$  & $0.4$  \\
 $mem_{\beta_{inc}}$ & $0.001$ & $0.001$  & $0.001$ & $0.001$  &  $0.001$  & $0.001$  \\
 $\gamma$ & $--$ & $0.98$  & $--$ & $0.98$  &  $--$  & $0.98$  \\
 $N_{mem}$ & $10^6$ & $10^6$  & $10^6$ & $10^6$ & $10^6$ & $10^6$  \\
 $N^{demo}_{mem}$ & $10^5$ & $10^5$ & $10^5$ & $10^5$ & $10^5$ & $10^5$ \\
 $N_{batch}$ & $2048$ & $2048$  & $2048$ & $2048$ & $2048$  & $2048$  \\
 \hline
\end{tabular}
\caption{The meta-parameter configurations used for HER.}
\label{table:configs_her}

\begin{tabular}{l|ll|ll|ll}
 & \multicolumn{2}{c}{\textbf{Reacher}} & \multicolumn{2}{|c}{\textbf{Half-Cheetah}} & \multicolumn{2}{|c}{\textbf{Walker}}\\
 & \textbf{Actor} & \textbf{Policy}  & \textbf{Actor} & \textbf{Policy}  & \textbf{Actor} & \textbf{Policy} \\
 \hline 
 $\phi(\cdot)$ & Tanh  &  Tanh &  Tanh &  Tanh &  Tanh & Tanh   \\
 Dims & $[512, 512]$ & $[512, 512]$ & $[512,512]$ & $[512,512]$ & $[512,512]$  & $[512,512]$  \\
 Rule & Adam & Adam & Adam & Adam & Adam & Adam   \\
 $\eta$ & $0.0001$ & $0.0003$  & $0.0005$ & $0.0004$ & $0.0005$  & $0.0002$  \\
 $\delta$ & $0.95$ & $0.95$  & $0.95$ & $0.95$ & $0.95$  & $0.95$  \\
 $low_{kl}$ & $5e-7$ & $5e-7$  & $5e-7$ & $5e-7$ & $5e-7$  & $5e-7$  \\
 $high_{kl}$ & $0.01$ & $0.01$  & $0.01$ & $0.01$ & $0.01$  & $0.01$  \\
 $\delta_0$ & $0.01$ & $0.01$  & $0.2$ & $0.2$ & $0.05$  & $0.05$  \\
 $k_{\delta}$& $50$ & $50$  & $50$ & $50$ & $50$  & $50$  \\
 $\gamma$ & $--$ & $0.991$  & $--$ & $0.991$  &  $--$  & $0.991$  \\
 $N_{mem}$ & $10^6$ & $10^6$  & $10^6$ & $10^6$ & $10^6$ & $10^6$  \\
 $N^{demo}_{mem}$ & $10^5$ & $10^5$ & $10^5$ & $10^5$ & $10^5$ & $10^5$ \\
 $N_{batch}$ & $2048$ & $2048$  & $2048$ & $2048$ & $2048$  & $2048$  \\
 \hline
\end{tabular}
\caption{The meta-parameter configurations used for LOGO.}
\label{table:configs_logo}

\end{center}
\end{table*}

\begin{table*}
\begin{center}

\begin{tabular}{l|lll|lll}
 & \multicolumn{3}{c}{\textbf{Hopper}} & \multicolumn{3}{|c}{\textbf{Swimmer}}\\
 & \textbf{Actor} & \textbf{Policy} & \textbf{Generator} & \textbf{Actor} & \textbf{Policy} & \textbf{Generator} \\
 \hline 
 $\phi(\cdot)$ & ReLU6  &  ReLU6 & ReLU6  &  ReLU6 & ReLU6  &  ReLU6 \\
 Dims & $[256, 256]$ & $[256, 256]$  & $[256,256]$ & $[256,256]$ & $[256,256]$ & $[256,256]$  \\
 Rule & Adam & Adam & Adam & Adam & Adam & Adam  \\
 $\eta$ & $0.0005$ & $0.0005$ & $0.0005$ & $0.0004$ & $0.0004$ & $0.0004$ \\
 $\tau$ & $0.06$ & $0.06$ & --  & $0.1$ & $0.1$ & -- \\
 $\gamma$ & -- & $0.99$ & -- &  -- & $0.99$ & --   \\
 $N_{mem}$ & $10^6$ & $10^6$ & $10^6$ & $10^6$ & $10^6$ & $10^6$  \\
 $N^{demo}_{mem}$ & $10^5$ & $10^5$ & $10^5$ & $10^5$ & $10^5$ & $10^5$ \\
 $N^{actor}_{mem}$ & $10^5$ & $10^5$ & $10^5$ & $10^5$ & $10^5$ & $10^5$ \\
 $N_{batch}$ & $256$ & $256$ & $128$ & $128$ & $256$ & $256$ \\
 \hline
\end{tabular}
\caption{The meta-parameter configurations used for ActPC.}
\label{table:configs_actpc_2}
\begin{tabular}{l|ll|ll}
 & \multicolumn{2}{c}{\textbf{Hopper}} & \multicolumn{2}{|c}{\textbf{Swimmer}} \\
 & \textbf{Actor} & \textbf{Critic}  & \textbf{Actor} & \textbf{Critic}  \\
 \hline 
 $\phi(\cdot)$ & ReLU  &  ReLU &  ReLU&  ReLU  \\
 Dims & $[512, 256]$ & $[512, 256]$ & $[512,256]$ & $[512,256]$ \\
 Rule & AdamW & Adam & Adam & Adam  \\
 $\eta$ & $0.0001$ & $0.0003$  & $0.0005$ & $0.0004$  \\
 $\tau$ & $0.005$ & $0.005$  & $0.005$ & $0.005$   \\
 $policy_{noise}$ & $--$ & $0.3$  & $--$ & $0.3$  \\
 $policy_{freq}$ & $--$ & $5$  & $--$ & $5$  \\
 $expl_{noise}$ & $0.1$ & $0.1$  & $0.1$ & $0.1$ \\
 $\gamma$ & $--$ & $0.991$  & $--$ & $0.992$   \\
 $N_{mem}$ & $10^6$ & $10^6$  & $10^6$ & $10^6$   \\
 $N^{demo}_{mem}$ & $10^5$ & $10^5$ & $10^5$ & $10^5$  \\
 $N_{batch}$ & $1024$ & $1024$  & $1024$ & $1024$  \\
 \hline
\end{tabular}
\caption{The meta-parameter configurations used for TD3.}
\label{table:configs_td3_1}

\begin{tabular}{l|ll|ll}
 & \multicolumn{2}{c}{\textbf{Hopper}} & \multicolumn{2}{|c}{\textbf{Swimmer}} \\
 & \textbf{Actor} & \textbf{Critic}  & \textbf{Actor} & \textbf{Critic}  \\
 \hline 
 $\phi(\cdot)$ & ReLU  &  ReLU &  ReLU&  ReLU  \\
 Dims & $[256, 256]$ & $[256, 256]$ & $[256,256]$ & $[256,256]$  \\
 Rule & AdamW & Adam & Adam & Adam  \\
 $\eta$ & $0.0001$ & $0.00003$  & $0.0005$ & $0.00004$  \\
 $\epsilon$ & $1.0$ & $1.0$  & $1.0$ & $1.0$   \\
 $\epsilon_{decay}$ & $0.9999$ & $0.9999$  & $0.9999$ & $0.9999$   \\
 $\epsilon_{min}$ & $0.32$ & $0.32$  & $0.32$ & $0.32$  \\
 $mem_{\alpha}$ & $0.6$ & $0.6$  & $0.6$ & $0.6$ \\
 $mem_{\epsilon}$ & $0.01$ & $0.01$  & $0.01$ & $0.01$   \\
 $mem_{\beta}$ & $0.4$ & $0.4$  & $0.4$ & $0.4$    \\
 $mem_{\beta_{inc}}$ & $0.001$ & $0.001$  & $0.001$ & $0.001$   \\
 $\gamma$ & $--$ & $0.98$  & $--$ & $0.98$   \\
 $N_{mem}$ & $10^6$ & $10^6$  & $10^6$ & $10^6$ \\
 $N^{demo}_{mem}$ & $10^5$ & $10^5$ & $10^5$ & $10^5$  \\
 $N_{batch}$ & $2048$ & $2048$  & $2048$ & $2048$  \\
 \hline
\end{tabular}
\caption{The meta-parameter configurations used for HER.}
\label{table:configs_her_1}

\begin{tabular}{l|ll|ll}
 & \multicolumn{2}{c}{\textbf{Hopper}} & \multicolumn{2}{|c}{\textbf{Swimmer}} \\
 & \textbf{Actor} & \textbf{Critic}  & \textbf{Actor} & \textbf{Critic}  \\
 \hline 
 $\phi(\cdot)$ & Tanh  &  Tanh &  Tanh &  Tanh \\
 Dims & $[512, 512]$ & $[512, 512]$ & $[512,512]$ & $[512,512]$  \\
 Rule & Adam & Adam & Adam & Adam  \\
 $\eta$ & $0.0001$ & $0.0003$  & $0.0005$ & $0.0004$  \\
 $\delta$ & $0.95$ & $0.95$  & $0.95$ & $0.95$  \\
 $low_{kl}$ & $5e-7$ & $5e-7$  & $5e-7$ & $5e-7$  \\
 $high_{kl}$ & $0.01$ & $0.01$  & $0.01$ & $0.01$  \\
 $\delta_0$ & $0.01$ & $0.01$  & $0.2$ & $0.2$  \\
 $k_{\delta}$& $50$ & $50$  & $50$ & $50$  \\
 $\gamma$ & $--$ & $0.991$  & $--$ & $0.991$    \\
 $N_{mem}$ & $10^6$ & $10^6$  & $10^6$ & $10^6$ \\
 $N^{demo}_{mem}$ & $10^5$ & $10^5$ & $10^5$ & $10^5$ = \\
 $N_{batch}$ & $2048$ & $2048$  & $2048$ & $2048$  \\
 \hline
\end{tabular}
\caption{The meta-parameter configurations used for Logo.}
\label{table:configs_logo_1}
\end{center}
\end{table*}

\section*{Hyper-parameter Configurations}

In Tables \ref{table:configs_actpc}-\ref{table:configs_logo}, we present the key hyper-parameter settings for ActPC as well as the baselines that we used in our experiments (models were tuned via a combination of controlled grid search and manual guidance informed from literature mediated by preliminary experimentation). 
$\phi(.)$ indicates choice of activation function for the hidden layers of either each NGC circuit (in the case of ActPC) or each MLP (in case of the baselines), ``Dims'' indicates the number of neurons in each layer of each circuit/MLP, $\eta$ is the learning rate of the optimization process (for a given ``Rule'' such as ``Adam'' or ``Adam'', also indicated in each table), $\tau$ indicates that Polyak averaging was used to update the corresponding network while $C$ indicates a hard frequency (number of transitions seen so far before the weights of the target network are set equal to its source network), $N_{mem}$ denotes the capacity of the uniform experience replay buffer while $N^{demo}_{mem}$ denotes the capacity of the demonstrate data buffer ($N^{actor}_{mem}$ is used for the actor self-imitation memory size in the case of ActPC), and $N_{batch}$ indicates the mini-batch sized used for sampling transitions to perform a synaptic update (or gradient in the case of MLPs/baselines). 

\begin{table*}[bt!]
\begin{center}
\begin{tabular}{l|lll|lll}
 & \multicolumn{3}{c}{\textbf{Block Lifting}} & \multicolumn{3}{|c}{\textbf{Can Pick-and-Place}}\\
 & \textbf{Actor} & \textbf{Policy} & \textbf{Generator} & \textbf{Actor} & \textbf{Policy} & \textbf{Generator} \\
 \hline 
 $\phi(\cdot)$ & ReLU6  &  ReLU6 & ReLU6  &  ReLU6 & ReLU6  &  ReLU6 \\
 Dims & $[256, 256]$ & $[256, 256]$  & $[256,256]$ & $[256,256]$ & $[256,256]$ & $[256,256]$  \\
 Rule & Adam & Adam & Adam & Adam & Adam & Adam  \\
 $\eta$ & $0.001$ & $0.001$ & $0.001$ & $0.001$ & $0.001$ & $0.001$ \\
 $C$ & $500$ & $500$ & --  & $500$ & $500$ & -- \\
 $\gamma$ & -- & $0.99$ & -- &  -- & $0.99$ & --   \\
 $N_{mem}$ & $10^6$ & $10^6$ & $10^6$ & $10^6$ & $10^6$ & $10^6$  \\
 $N^{demo}_{mem}$ & $1.2 * 10^5$ & $1.2 * 10^5$ & $1.2 * 10^5$ & $1.2 * 10^5$ & $1.2 * 10^5$ & $1.2 * 10^5$  \\
 $N^{actor}_{mem}$ & $2*10^5$ & $2*10^5$ & $2*10^5$ & $2*10^5$ & $2*10^5$ & $2*10^5$ \\
 $N_{batch}$ & $256$ & $256$ & $256$ & $256$ & $256$ & $256$ \\
 \hline
\end{tabular}
\caption{The robosuite meta-parameter configurations used for ActPC.}
\label{table:configs_actpc_1}
\begin{tabular}{l|ll|ll}
 & \multicolumn{2}{c}{\textbf{Block Lifting}} & \multicolumn{2}{|c}{\textbf{Can Pick-and-Place}} \\
 & \textbf{Actor} & \textbf{Critic}  & \textbf{Actor} & \textbf{Critic}  \\
 \hline 
 $\phi(\cdot)$ & ReLU  &  ReLU &  ReLU&  ReLU   \\
 Dims & $[512, 256]$ & $[512,512]$ & $[512,256]$ & $[512,256]$   \\
 Rule & Adam & Adam & Adam & Adam  \\
 $\eta$ & $0.0003$ & $0.0005$  & $0.0003$ & $0.00035$  \\
 $\tau$ & $0.02$ & $0.05$  & $0.05$ & $0.05$   \\
 $\gamma$ & -- & $0.98$  & -- & $0.989$  \\
 $N_{mem}$ & $10^6$ & $10^6$  & $10^6$ & $10^6$  \\
 $N^{demo}_{mem}$ & $1.2 * 10^5$ & $1.2 * 10^5$ & $1.2 * 10^5$ & $1.2 * 10^5$  \\
 $N_{batch}$ & $512$ & $512$  & $512$ & $512$  \\
 \hline
\end{tabular}
\caption{The robosuite meta-parameter configurations used for DDPG-Demo.}
\label{table:configs_ddpg_demo}
\end{center}
\end{table*}

\end{document}